\pdfoutput=1

\documentclass[11pt]{article}

\usepackage{acl}

\usepackage{times}
\usepackage{latexsym}
\usepackage{subfigure}
\usepackage{amssymb}
\usepackage{multirow}
\usepackage{subcaption}  
\usepackage{amsmath} 

\usepackage{booktabs}

\usepackage[T1]{fontenc}

\usepackage[utf8]{inputenc}

\usepackage{microtype}

\usepackage{inconsolata}

\usepackage{graphicx}

\title{A Representation Level Analysis of NMT Model Robustness to Grammatical Errors}

\author{Abderrahmane Issam \\ {\bf Yusuf Can Semerci} \\ {\bf Jan Scholtes} \\ {\bf Gerasimos Spanakis} \\
        Department of Advanced Computing Sciences \\ 
        Maastricht University \\ 
        \small{\texttt{\{abderrahmane.issam, y.semerci, j.scholtes, jerry.spanakis\}@maastrichtuniversity.nl}}}

\begin{document}
\maketitle
\begin{abstract}

Understanding robustness is essential for building reliable NLP systems. Unfortunately, in the context of machine translation, previous work mainly focused on documenting robustness failures or improving robustness. In contrast, we study robustness from a model representation perspective by looking at internal model representations of ungrammatical inputs and how they evolve through model layers. For this purpose, we perform Grammatical Error Detection (GED) probing and representational similarity analysis. Our findings indicate that the encoder first detects the grammatical error, then corrects it by moving its representation toward the correct form. To understand what contributes to this process, we turn to the attention mechanism where we identify what we term \textit{Robustness Heads}. We find that \textit{Robustness Heads} attend to interpretable linguistic units when responding to grammatical errors, and that when we fine-tune models for robustness, they tend to rely more on \textit{Robustness Heads} for updating the ungrammatical word representation. \footnote{\texttt{Our code: \url{https://github.com/issam9/nmt-robustness-analysis}}.}

\end{abstract}

\section{Introduction}

Neural Machine Translation (NMT) has seen great success,  especially since the introduction of the Transformer architecture \cite{vaswaniattention}. Recent advances in NMT introduced models that can translate between over 200 languages \cite{nllbteam2022language}. While this achievement is impressive and drives the deployment in real-world scenarios, evaluating and understanding NMT robustness remains essential for building reliable NMT systems.

Early works have focused on documenting the robustness failures of NMT models, or improving their robustness \cite{napoles-etal-2016-effect, khayrallah-koehn-2018-impact, belinkov2018synthetic, anastasopoulos-2019-analysis, jayanthi-pratapa-2021-study}. However, there has been limited analysis of model representations in response to noise. Therefore, our goal in this work is to fill this gap by analyzing robustness from a representation perspective. 

Our hypothesis is that the encoder detects and corrects the representation of the ungrammatical word by moving its representation toward the correct form. To study the detection part, we use GED probing to evaluate how the accuracy of detecting the ungrammatical word changes through the encoder layers. For the correction part, we measure the representation distance between the ungrammatical word and its correct grammatical form. We find that generally the probing performance increases in the first half layers of the model, then plateaus or decreases, while the representation distance on the overall decreases along model layers. 

To understand what contributes to correcting the representations, we turn to the attention mechanism due to its crucial contribution to transformer model performance. We identify what we term \textit{Robustness Heads}, which are attention heads that contribute to moving the ungrammatical word's representation toward its correct form. We find that after fine-tuning models on ungrammatical sentences, and thus, making them more robust, they learn to rely more on \textit{Robustness Heads} for updating the ungrammatical word's representation especially in deeper layers where we hypothesize that the correction is happening.

Our work addresses the following research questions and makes the following contributions:

\noindent
\textbf{RQ1:} \textit{How do models represent and handle grammatical errors to achieve robustness?} NMT encoders inherently implement a Grammatical Error Correction (GEC) setup in which they first detect the ungrammatical word, then correct it by moving its representation toward the grammatical form.

\noindent
\textbf{RQ2:} \textit{How does grammatical error representation differ across models and languages?} Models respond similarly to grammatical errors however we find differences across languages which we attribute to their linguistic differences.

\noindent
\textbf{RQ3:} \textit{How does fine-tuning for robustness influences models to achieve improved robustness?} Fine-tuned models exhibit similar behavior to their base models but tend to rely more on \textit{Robustness Heads} for handling grammatical errors and enhancing their resilience to noisy inputs.

\section{Related Work}

There have been multiple works analyzing transformer models representations for different purposes. Probing has been explored for understanding information that is encoded in pre-trained model representations \cite{belinkov-etal-2017-evaluating, ettinger-2020-bert, belinkov-etal-2020-linguistic, liu-etal-2021-complementarity-pre, davis-etal-2022-probing}, as well as to understand the effect of fine-tuning \cite{mosbach-etal-2020-interplay, durrani-etal-2021-transfer, zhou-srikumar-2022-closer, nadir-etal-2023-discovering}, and contextual embeddings \cite{tenney2018what, klafka-ettinger-2020-spying}. The attention module, given its importance in the success of the transformer architecture, has also been explored for interpreting transformer models \cite{raganato-tiedemann-2018-analysis, clark-etal-2019-bert, voita-etal-2019-bottom, zhang-etal-2023-closer}. Other works have focused on understanding and comparing between different models through the lens of similarity analysis \cite{kudugunta-etal-2019-investigating, wu-etal-2020-similarity, vazquez-etal-2021-differences}. In our work, we combine all these techniques to understand NMT models robustness to grammatical errors.

Fine-tuning on downstream tasks leads to changes in representations that might not be favorable (e.g. catastrophic forgetting), and thus has been an active area of analysis \cite{merchant-etal-2020-happens, durrani-etal-2021-transfer, phang-etal-2021-fine, zhou-srikumar-2022-closer, durrani-2023-discovering, neerudu-etal-2023-robustness}. Our work focuses on understanding the effects of fine-tuning for robustness to grammatical errors in the context of NMT, where we look at both model representations and downstream NMT performance.

Robustness is of critical importance for NLP models. Early works have explored improving robustness to different types (e.g. User Generated Data, Automatic Speech Recognition, Non native speakers, ...) of noise by using synthetic data \cite{anastasopoulos-etal-2019-neural,zhou-etal-2019-improving, karpukhin-etal-2019-training, salesky-etal-2019-fluent, jayanthi-pratapa-2021-study, wang-etal-2021-secoco-self, zhao-calapodescu-2022-multimodal}. Recent works show that even Large Language Models (LLMs) witness performance degradation when confronted with synthetic noise \cite{chen2024nlperturbatorstudyingrobustnesscode,zhu2024promptrobustevaluatingrobustnesslarge}. Multiple techniques were introduced to deal with noise or adversarial attacks by pushing noisy or adversarial samples representations to be similar to those of the original samples \cite{xu-etal-2021-addressing, passban-etal-2021-revisiting-robust, yang2022robust, wang-etal-2023-robustness} and our experiments show that NMT encoders do this inherently when trained on synthetic noise. While analyzing the robustness of NMT transformer models has been an area of exploration \cite{napoles-etal-2016-effect, khayrallah-koehn-2018-impact, belinkov2018synthetic, anastasopoulos-2019-analysis, jayanthi-pratapa-2021-study}, the focus was more on documenting model failures under noise or adversarial attacks rather than analyzing the model internals, therefore, our work is an attempt to fill this gap. In computer vision, however, \citet{cianfarani2022understanding} explored representation similarity  to understand adversarially trained Deep Neural Networks (DNNs) and compare them to non-robust DNNs, while we focus on understanding the effect of fine-tuning NMT transformer models on synthetic grammatical errors, and we compare them against their base model as a less robust model, as well as against a domain adapted model. In addition, we probe the linguistic features encoded in the representations and we compare the models on the basis of their attention mechanism.

\section{Methodology}
Our analysis starts by introducing grammatical errors in the dataset, which we describe in \S\ref{synthetic-errors}. Subsequently, in \S\ref{finetuning_for_robustness} we describe how we fine-tune models for robustness to analyze the effects of fine-tuning. Finally, in \S\ref{model_analysis} we present the methods of our analysis, namely: GED probing, representation similarity, and \textit{Robustness Heads}, where we describe our method for finding \textit{Robustness Heads}, and analyzing their attention to POS tags.
\subsection{Synthetic Grammatical Errors}
\label{synthetic-errors}
To provide a representation level analysis of robustness, it is crucial to have granular control over grammatical errors. We achieve this by introducing synthetic grammatical errors into clean sentences. We focus on three types of grammatical errors that are common in non-native speaker language \cite{izumi-etal-2004-overview, napoles-etal-2016-effect, anastasopoulos-2019-analysis}, and we create an adversarial copy of the dataset for each type, where we insert one error per sentence when possible. We focus on grammatical errors with clear linguistic functions to be able to link our analysis to the linguistic features of the source language. However, to validate the generalizability of our analysis to different types of errors as well as more than one error per sentence, we use MORPHEUS \cite{tan-etal-2020-morphin} as a black-box adversarial attack that greedily introduces inflectional errors to minimize the BLEU score.

We follow the implementation of \cite{anastasopoulos-etal-2019-neural} to introduce article (\textit{Article}), preposition (\textit{Prep}) and noun number (\textit{Nounnum}) replacement errors in the dataset. For each sentence $X = \{w_1, w_2, w_3, ..., w_n\}$ in the dataset $D$, we introduce one of the three grammatical errors in $X$ when possible to get $\widetilde{X}$ = $\{w_1, w_2, \widetilde{w_3}, ..., w_n\}$, where $\widetilde{w_3}$ is the noisy word that was sampled to replace $w_3$. Therefore, $\widetilde{w_3}$ represents the ungrammatical word, and $w_3$ is its target grammatical form. The result is an adversarial dataset $\widetilde{D}$ for each error type.

\subsection{Fine-tuning for Robustness}
\label{finetuning_for_robustness}

Our analysis focuses on four well established NMT models, namely: OPUS-MT, M2M100, MBART and NLLB. We fine-tune these models on the adversarial dataset of one of the error types, and since this leads to improving their robustness to the error type, we also analyze the representations of fine-tuned models. To separate the effect of robustness from domain adaptation, we compare against a version of the model that is fine-tuned on the clean version of the data. We refer to these models as Base, Noise-Finetuned, and Clean-Finetuned respectively. However, we only fine-tune the encoder given that it is the source-side representation engine of the model. We justify this focus in Appendix \ref{freezing} where we find that fine-tuning only the encoder achieves similar robustness to fine-tuning the full model, which is not the case when fine-tuning only the decoder or the cross attention. 

\subsection{Model Analysis}
\label{model_analysis}
\subsubsection{GED Probing}

Following \cite{davis-etal-2022-probing}, we perform GED probing to understand how the detection of ungrammatical words changes through the encoder layers. We train linear probes on the word representation of each encoder layer to predict whether the word is grammatically correct or not. Similarly to previous work \cite{liu-etal-2019-linguistic, davis-etal-2022-probing}, we take the representation of the last subword as the word representation when a word is split into subwords. 

\subsubsection{Representation Similarity}
To study how encoders affect the representation of the ungrammatical word toward its correct form, we measure the distance between the ungrammatical word and its target grammatical form in each of the encoder layers. We use Centered Kernel Alignment (CKA) \cite{simon-cka-2019} to measure the distance as $1-CKA(\widetilde{W}, W)$, where $\widetilde{W} \in \mathbb{R}^{N \times d}$ are ungrammatical word representations, $W \in \mathbb{R}^{N \times d}$ are their target grammatical word representations, $N$ is the number of data points, and $d$ is the model hidden dimension. We use CKA because compared to other similarity methods, it does not require the number of data points to be considerably higher than the representation dimension \cite{simon-cka-2019}. We note that CKA outputs a similarity score between 0 and 1.

\subsubsection{Robustness Heads}
\subsubsection{From Influential Heads to Robustness Heads}
\citet{voita-etal-2019-bottom} measured the amount of influence of a token $w_i$ on another token $w_j$ as the distance between $w_j$'s representation before and after $w_i$ was masked. We apply the same method but to measure head influence instead. We collect the word representation after masking one head at a time, then we compute the CKA distance to the original word representation from the same layer. In our hypothesis, masking an attention head that has the most influence on the word's representation will lead to the highest distance, therefore, we term this \textit{Influential Heads}. Formally, for a layer $l$, we mask each head $h_i \in \{1, 2, ..., H\}$ in layer $l-1$, and we take the representation of the word $w$ to compute $1-CKA(w_{h_i}, w)$, where $H$ is the number of heads, $w_{h_i}$ is the word representation after masking the head $h_i$, and $w$ is the original word representation. 

Since we are more interested in understanding which attention heads contribute to correcting the ungrammatical word's representation, we are led to introduce what we term \textit{Robustness Heads}, which we define as heads that influence the ungrammatical word's representation toward its grammatical form. This requires a simple redefinition of \textit{Influential Heads}, where instead of computing the distance to the original word itself, we compute the distance from the noisy word representation to the representation of its clean form. Formally, instead of computing $1-CKA(w_{h_i}, w)$ we compute $1-CKA(\widetilde{w_{h_i}}, w)$, where $\widetilde{w_{h_i}}$ is the representation of the ungrammatical word $\widetilde{w}$ after head $h_i$ is masked.

\subsubsection{Attention to POS Tags}
The attention mechanism offers a straightforward way to interpret NLP models, based on the assumption that, like human attention, models focus on parts of the sentence that they find important for making a prediction. This assumption combined with the granularity and clear linguistic functions of the grammatical errors we introduce, offer a way to linguistically analyze the attention of \textit{Robustness Heads}, which we achieve by inspecting their attention to Part Of Speech (POS) Tags. We collect the attention scores directed from the ungrammatical tokens to the other tokens in the sentence, then group the attention scores over words, following \cite{clark-etal-2019-bert}. When the noisy word is split into tokens, we take the mean of the scores. Conversely, when the word it is attending to is split into tokens, we take their sum. This preserves the property that word level attention scores sum to 1. Following this transformation, we use Spacy \footnote{\texttt{\url{https://spacy.io/}}} to label each word in the sentence with its POS tag and collect the attention scores that each POS tag has received.

\section{Experimental Setup}
\subsection{Data}
We use the Europarl-ST \cite{europarl-st} dataset which contains official speech, transcriptions and translations of European Parliament debates of multiple European languages. For this work, we only use the transcriptions and translations of 5 directions: En-Es, En-De, En-It, En-Nl and Fr-Es (Dataset splits and sizes are in \S\ref{data_splits_appendix}). Introducing grammatical errors and interpreting the results of our analysis required understanding of the source languages, which limited the choice of language directions. We were also limited by the performance of models (e.g. MBART scores 51.18 COMET on Fr-Nl) and availability of NLP resources for introducing grammatical errors. Nevertheless, there is a high number of non-native speakers of both English and French, which makes them relevant for our analysis.

After we introduce synthetic errors into the dataset, we sample 30\% of the train set for training the GED probes and we use the rest for fine-tuning the models. When sampling, we make sure that the error labels are balanced between the two subsets. Since we seek to experiment with fine-tuning on clean data as well, we keep the clean version of the fine-tuning subset. This results in clean and noisy fine-tuning subsets for the 3 grammatical errors that we introduce, which we use to fine-tune Clean-Finetuned and Noise-Finetuned models of each error type. While for GED probing, we only use the noisy version to probe the models on each error type.

\subsection{Synthetic Errors}
\label{grammatical_errors_appendix}

\noindent
\textbf{\textit{Nounnum}}: We find nouns in the sentence then sample one of them and change its number from plural to singular or the opposite depending on its actual number. For English, we use Berkeley parser \cite{petrov-etal-2006-learning} to identify nouns and their number, and for French we use Spacy-lefff \footnote{\texttt{\url{https://github.com/sammous/spacy-lefff}}}.

\noindent
\textbf{\textit{Article} and \textit{Prep}}: We find articles or prepositions in the sentence with string matching, and sample one of them uniformly, then sample a replacement based on statistics from CONLL-14 Shared Task on GEC \cite{ng-etal-2014-conll} dataset in the case of English, and uniformly in the case of French. The list of articles and prepositions is provided in Appendix \ref{articles_prepositions}.

\textbf{\textit{Morpheus}}: MORPHEUS \cite{tan-etal-2020-morphin} is a black-box adversarial attack that greedily introduces inflectional errors to minimize or maximize a target metric, which in our case is the BLEU score. Similarly to the original work, we only inflect nouns, verbs and adjectives, and we restrict the possible inflections to the original POS tag to preserve the meaning of the sentence. We use Spacy and Spacy-lefff for tokenization and POS tagging of English and French sentences respectively, and we use LemmInflect \footnote{\texttt{\url{https://github.com/bjascob/LemmInflect}}} to find inflections for English similarly to the original work, and Inflecteur \footnote{\texttt{\url{https://github.com/Achuttarsing/inflecteur}}} to find inflections for French. For further details, we refer the reader to the original work \cite{tan-etal-2020-morphin}.

\subsection{Models}
We run experiments on multilingual machine translation models. More specifically, we analyzed three models: Multilingual translation version of \href{https://huggingface.co/facebook/mbart-large-50-many-to-many-mmt}{MBART} \cite{tang2020multilingualtranslationextensiblemultilingual}, \href{https://huggingface.co/facebook/m2m100\_418M}{M2M100} \cite{fan2021m2m100} and \href{https://huggingface.co/facebook/nllb-200-distilled-600M}{NLLB} \cite{nllbteam2022language}. We also experimented with bilingual models of OPUS-MT \cite{tiedemann-thottingal-2020-opus} for \href{https://huggingface.co/Helsinki-NLP/opus-mt-en-es}{En-Es}, \href{https://huggingface.co/Helsinki-NLP/opus-mt-en-de}{En-De}, \href{https://huggingface.co/Helsinki-NLP/opus-mt-en-it}{En-It}, \href{https://huggingface.co/Helsinki-NLP/opus-mt-en-nl}{En-Nl} and \href{https://huggingface.co/Helsinki-NLP/opus-mt-fr-es}{Fr-Es}. Each of these models has a different level of multilinguality, where OPUS-MT models support a single direction, while MBART, M2M100 and NLLB support many-to-many translation between 50 languages, 100 languages, and over 200 languages respectively. 

\subsection{Fine-tuning}
\label{finetuning_details}
We fine-tune the models using HuggingFace transformers library \footnote{\texttt{\url{https://github.com/huggingface/transformers}}}. We use AdamW optimizer \cite{loshchilov2018decoupled} with a learning rate of 5e-05 and a batch size of 64. We save the best model based on development set BLEU score during a maximum of 5000 steps. For validation, we use the clean and noisy development sets when fine-tuning on the clean and noisy subsets respectively. We fine-tune the models with a frozen decoder unless it is mentioned otherwise.

\subsection{GED Probe Training}
\label{probing_details}
We train single layer probes using the Pytorch framework. we use Adam optimizer with a learning rate of 1e-03 and a weight decay of 1e-04. We train with a batch size of 32 for 50 epochs with a dropout of 0.1 and a patience of 10 epochs for early stopping based on validation F1 score.

\subsection{Evaluation} 
We evaluate translation on the Europarl-ST test sets. For each grammatical error, we report results on both clean and noisy test sets. We also report the difference in performance between clean and noisy. As metrics, we use COMET \cite{rei-etal-2020-comet}, BLEU \cite{bleu-2002} and ChrF \cite{popovic-2015-chrf}. COMET is a neural based metric that was shown to be more aligned with human judgments \cite{freitag-etal-2022-results}. We use the reference-based model wmt22-comet-da \footnote{\texttt{\url{https://huggingface.co/Unbabel/wmt22-comet-da}}}, and we report BLEU and ChrF results in our repository. 

For GED probing, we evaluate probes of each model on the noisy version of the test set of each grammatical error using the F1 score.

\section{Results and Analysis}
\subsection{Fine-tuning}
In Table \ref{clean_vs_noisy} we show the COMET scores of Base, Clean-Finetuned and Noise-Finetuned models on clean and noisy test sets and their difference ($\Delta$) for En-Es (and Table \ref{table:finetuning_results_appendix} for the other language directions). We can see that even in our simple setup, where we insert one error per sentence, we still see a significant drop in performance in Base models (0.66 at minimum) \cite{kocmi-etal-2024-navigating}. Furthermore, fine-tuning on clean or noisy data leads to better results but only fine-tuning on noisy data leads to improving robustness, which is seen in the reduced difference in COMET (e.g. from 0.74 to 0.01 for NLLB on \textit{Article} errors). Surprisingly, fine-tuning on grammatical errors doesn't affect performance on clean data, and can even lead to better results compared to fine-tuning on clean data (e.g. 77.51 compared to 77.14 for M2M100 on \textit{Prep} errors). This suggests that fine-tuning on grammatical errors has a regularization effect. Across error types, we see that \textit{Prep} errors lead to the most significant drop, and that even after fine-tuning, the drop in performance is still significant (e.g. up to 0.29 for NLLB). We note that Clean-Finetuned models performance on clean data is different across errors because the clean train subsets were sampled to match the noisy subsets which are different because of the error distribution.

In this sub-section, we established that grammatical errors lead to a significant drop in performance, and that fine-tuning on them leads to increased robustness, therefore in the next sub-section we proceed to analyze Base, Noise-Finetuned and Clean-Finetuned models representations when responding to the grammatical errors. Furthermore, in Appendix \ref{morpheus_results} we provide the results on \textit{Morpheus}, which confirms the generalizability of our analysis to other error types and to sentences containing more than one error per sentence.

\begin{table*}[ht]
\scriptsize
\centering
\begin{tabular}{llccccccccc}
\toprule
\textbf{Direction} & \textbf{Model} & \multicolumn{3}{c}{\textbf{Article}} & \multicolumn{3}{c}{\textbf{Nounnum}} & \multicolumn{3}{c}{\textbf{Prep}} \\
 &  & \textbf{Clean} & \textbf{Noisy} & \textbf{$\Delta$} & \textbf{Clean} & \textbf{Noisy} & \textbf{$\Delta$} & \textbf{Clean} & \textbf{Noisy} & \textbf{$\Delta$} \\
\midrule
\multirow{13}{*}{En-Es}       & opus-mt-base  & 78.72   & 77.71   & 1.0        & 78.72   & 77.74   & 0.97       & 78.72 & 77.67 & 1.05       \\
& opus-mt-clean & 78.88   & 78.05   & 0.84       & 78.97   & 78.12   & 0.85       & 78.93 & 78.05 & 0.88       \\
& opus-mt-noise & \textbf{78.94}   & \textbf{78.89}   & \textbf{0.06}       & \textbf{78.99}   & \textbf{78.82}   & \textbf{0.17}       & \textbf{79.06} & \textbf{78.83} & \textbf{0.23} \\
\cmidrule{2-11}
& m2m100-base   & 75.99   & 74.84   & 1.15       & 75.99   & 75.12   & 0.88       & 75.99 & 74.63 & 1.36       \\
& m2m100-clean  & 77.39   & 76.4    & 0.99       & 77.28   & 76.22   & 1.07       & 77.14 & 76.14 & 1.0        \\
& m2m100-noise  & \textbf{77.57}   & \textbf{77.54}   & \textbf{0.03}       & \textbf{77.58}   & \textbf{77.5}    & \textbf{0.08}       & \textbf{77.51} & \textbf{77.23} & \textbf{0.28}       \\
\cmidrule{2-11}
& mbart-base    & 78.04   & 77.23   & 0.81       & 78.04   & 77.25   & 0.79       & 78.04 & 77.24 & 0.79       \\
& mbart-clean   & 78.51   & 77.77   & 0.74       & \textbf{78.64}   & 77.74   & 0.9        & 78.49 & 77.73 & 0.76       \\
& mbart-noise   & \textbf{78.61}   & \textbf{78.58}   & \textbf{0.04}       & 78.57   & \textbf{78.51}   & \textbf{0.06}       & \textbf{78.58} & \textbf{78.38} & \textbf{0.2}        \\
\cmidrule{2-11}
& nllb-base     & 78.34   & 77.6    & 0.74       & 78.34   & 77.69   & 0.66       & 78.34 & 77.52 & 0.83       \\
& nllb-clean    & \textbf{78.82}   & 78.14   & 0.68       & \textbf{78.85}   & 78.16   & 0.69       & 78.81 & 78.21 & 0.6        \\
& nllb-noise    & 78.81   & \textbf{78.8}    & \textbf{0.01}       & 78.78   & \textbf{78.73}   & \textbf{0.06}       & \textbf{78.94} & \textbf{78.66} & \textbf{0.29}   
\\
\bottomrule
\end{tabular}
\caption{COMET scores on En-Es. The Base model is the original model, Clean is fine-tuned on the clean version of the data, and Noise is fine-tuned on the noisy version (with the same noise as the one they are tested on). We present the performance on the clean and noisy test sets and their difference ($\Delta$).}
\label{clean_vs_noisy}
\end{table*}

\subsection{GED Probing}
\label{ged_probing}

\begin{figure}[ht]
    \centering
    \includegraphics[width=0.47\textwidth]{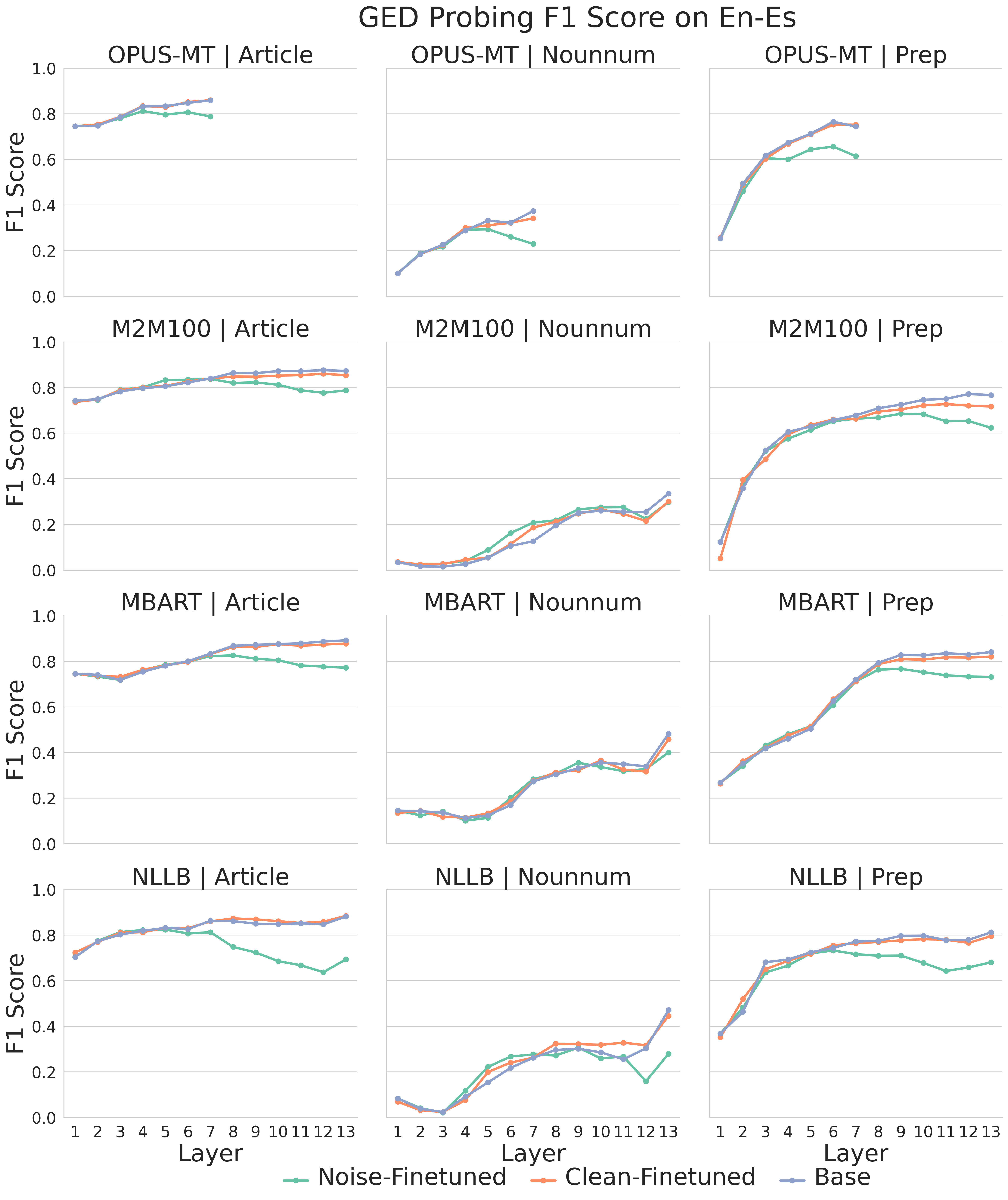}
    \caption{GED probing performance of Noise-Finetuned, Clean-Finetuned and Base models on En-Es. GED probing performance of Noise-Finetuned models witnesses a degradation in deeper layers.}
    \label{fig:ged_probing}
\end{figure}

Figure \ref{fig:ged_probing} (and Figure \ref{fig:ged_probing_appendix} for the other language directions) shows that the GED probing performance improves during roughly the first half layers of the model, then generally plateaus in the second half for Base and Clean-Finetuned but decreases for Noise-Finetuned models. Additionally, across Base models the representation of errors achieves closely similar GED F1 score on each layer, while across errors, the F1 scores are different. For example, the GED F1 score of \textit{Nounnum} errors at layer 1 is almost 0., while that of \textit{Article} errors is 0.7. When looking at French as a different source language, Figure \ref{fig:ged_probing_fr_es} shows that French error detection is represented differently, especially that of \textit{Nounnum} errors where the maximum F1 score on En-Es is 0.48, while on Fr-Es it is 0.84. This might be explained by the fact that in French, the noun number is indicated by adjectives and articles, while it is not the case for English.

Fine-tuning negatively affects the GED probing performance in deeper layers, where our GEC setup hypothesis suggests that correction is happening. Noise-Finetuned models maintain their ability to detect errors in lower layers, then correct them in deeper layers, which makes the GED probing accuracy more challenging because the representation is corrected. In the next sub-section, we further support this correction hypothesis.

\subsection{Representation Distance}
\label{representation_distance}
\begin{figure}[ht]
    \centering
    \includegraphics[width=0.47\textwidth]{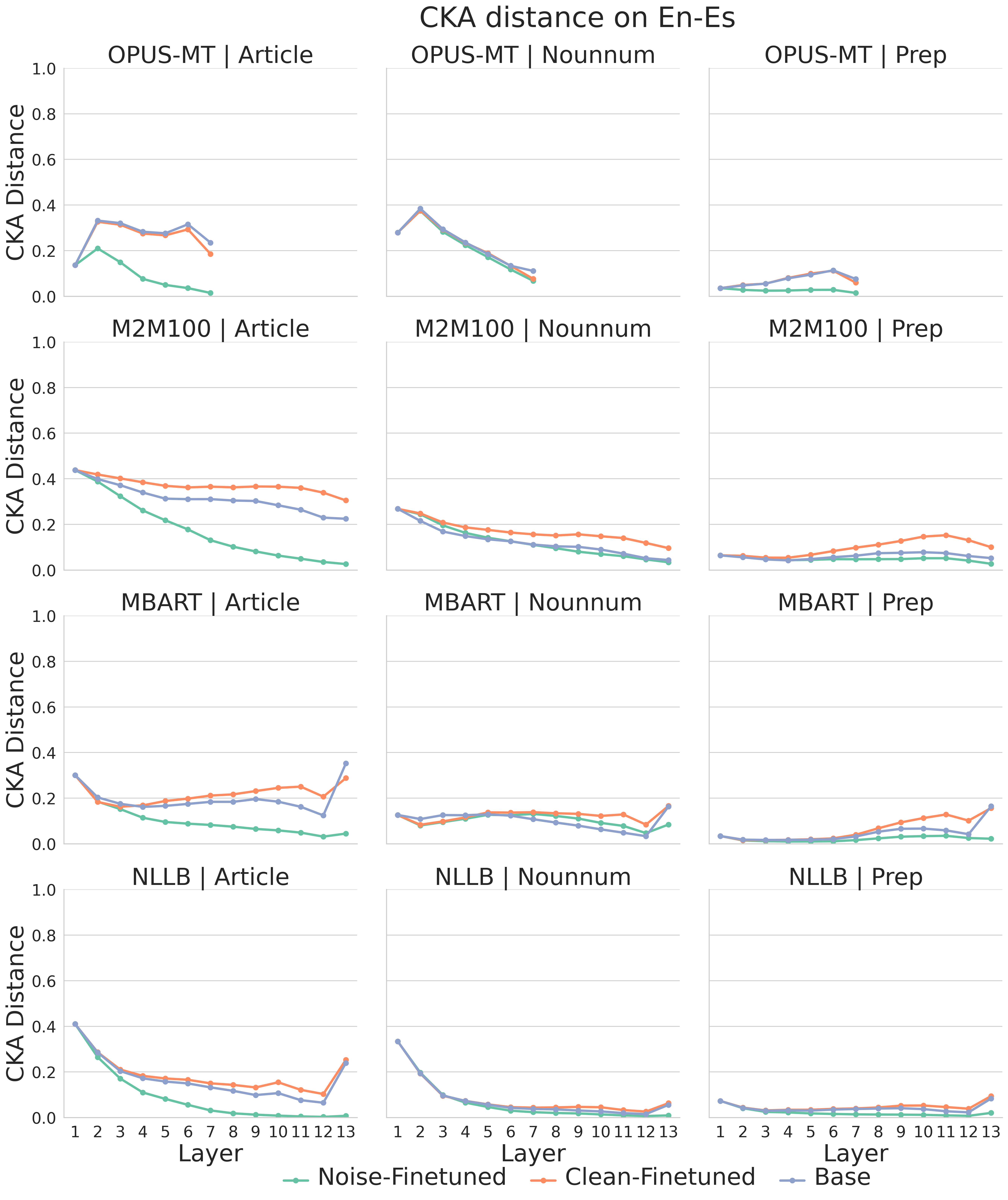}
    \caption{CKA distance of clean and noise word representations across models and errors on En-Es. Noise-Finetuned models drive the representation of the noisy word to be more similar to the clean word.}
    \label{fig:cka_distance}
\end{figure}

Figure \ref{fig:cka_distance} (and Figure \ref{fig:cka_distance_appendix} for other language directions) shows the representation distance between the ungrammatical or noisy word and its target clean word at each layer of the encoder. Generally, for Base and Clean-Finetuned models the CKA distance decreases from one layer to the next except for \textit{Prep} errors where the distance decreases then increases in deeper layers. This can be explained by the fact that both words have the same linguistic function (nouns, articles or prepositions), and because they share the same context which leads their representation to move closer as the encoder integrates context into it. On the other hand, Noise-Finetuned models exhibit similar behavior to their Base model but they learn to drive the representation to be closer (almost 0. CKA distance in most cases), this means that the similarity is driven by robustness as well, where models correct ungrammatical words by pushing their representation toward their grammatical form. Combined with the GED probing results, this supports our hypothesis that the encoder detects and corrects grammatical errors to achieve robustness. In the next sub-section, we analyze \textit{Robustness Heads} to explain this behavior in Noise-Finetuned Models.

\subsection{Robustness Heads}
\subsubsection{Attention to POS Tags}
\label{attention_pos_tags}

\begin{figure*}[ht]
    \centering
    \includegraphics[width=0.95\textwidth,height=0.42\textheight]
    {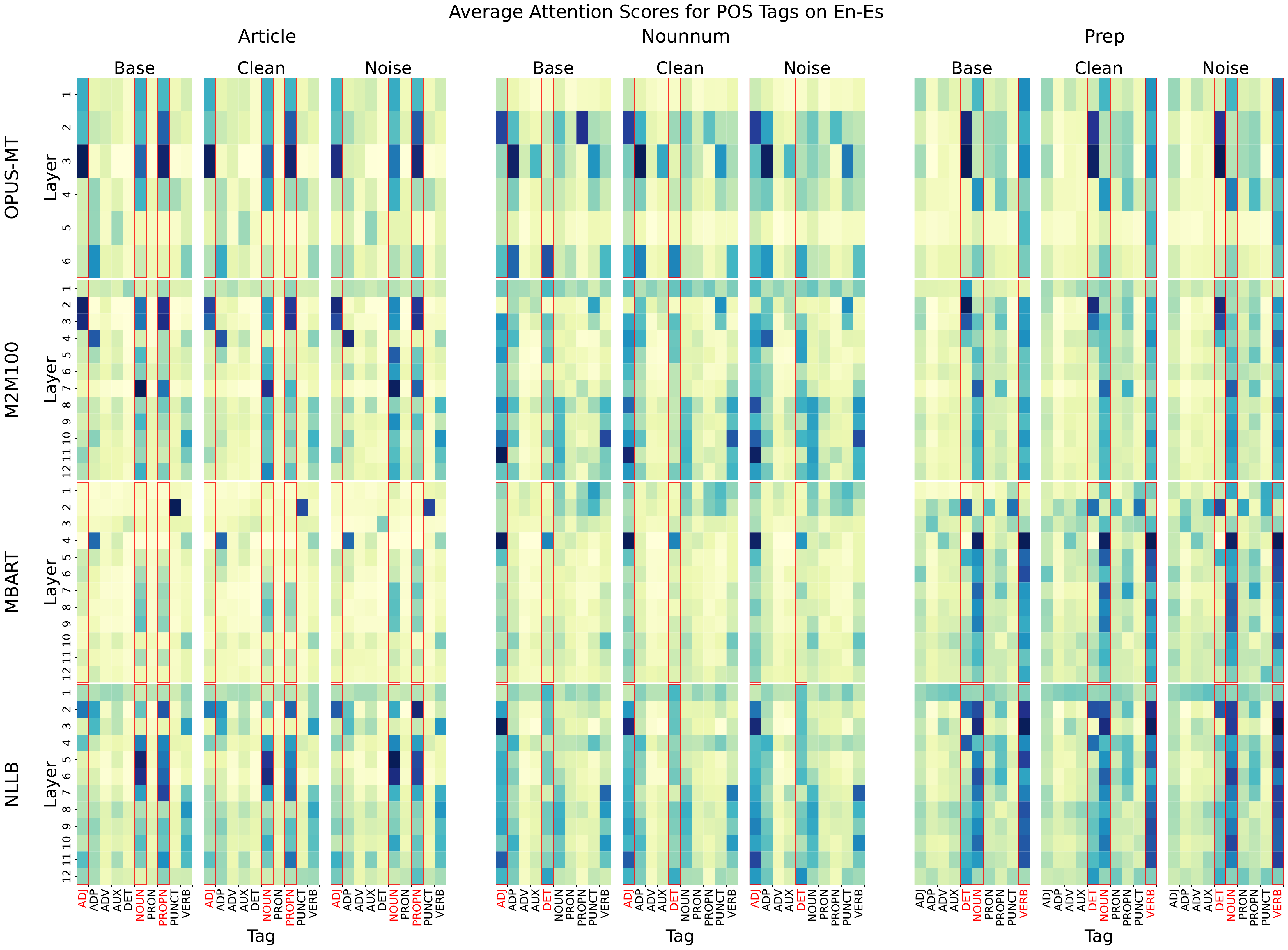} 
    
    \caption[\textit{Robustness Heads} attention to the 10 most common POS tags in the test set on En-Es. The scale of attention is relative to each base model and error. We highlight POS tags that are attended to the most across models.]{\textit{Robustness Heads} attention to the 10 most common POS tags in the test set on En-Es. The scale of attention is relative to each base model and error. We highlight POS tags that are attended to the most across models.}
    \label{fig:attention_heatmap}
\end{figure*}

Figure \ref{fig:attention_heatmap} (and Figures \ref{fig:attention_pos_fr_es}, \ref{fig:attention_pos_en_de}, \ref{fig:attention_pos_en_it}, \ref{fig:attention_pos_en_nl} for other language directions) presents the average attention scores of each POS tag. The scale of attention is relative to each base model and error type. For clarity reasons, we only show the attention scores over the 10 most common POS tags in the dataset. We keep the tags as they are named in Spacy, but their full names are presented in Appendix \ref{pos_tags_list}. The figure shows that generally, the attention of \textit{Robustness Heads} to POS tags is concentrated in the early layers which is related to how early work have found that lower layers are better at POS tagging \cite{belinkov-etal-2017-evaluating}. Furthermore, \textit{Robustness Heads} attend to words in the sentence that can help identify or correct the grammatical error in question. Although the attention is distributed differently across models, they still attend to similar POS tags when responding to the same error. However, when comparing between English and French, models attend to different POS tags, which can be explained by their linguistic differences. If we look at \textit{Article} errors, English models primarily focus their attention on adjectives, nouns, and proper nouns, while their French counterparts primarily focus on nouns. This difference can be attributed to the ordering of adjectives in each language. In English, adjectives mostly precede nouns, while the opposite is true for French. This means that adjectives in English have more direct influence on articles, especially when making the choice between "a" and "an" (they depend on whether the next word starts with a consonant or a vowel). Another fact that could lead the model to focus less on adjectives in French, is that certain adjectives do not follow their noun in gender or number. When examining \textit{Nounnum} errors, models exhibit a common pattern of focusing primarily on adjectives and determiners, however, attention to determiners in English is notably lower compared to French, which can be explained by the fact that in French, the noun number is indicated by articles. Finally, for \textit{Prep} errors, the models focus mainly on verbs, nouns, and determiners. The three definitely can affect the choice of prepositions; although it is not clear with determiners, we note that some prepositions are more common with definite vs. indefinite determiners, such as "on" and "the". 
\begin{figure}[ht!]
    \centering
    \includegraphics[width=0.47\textwidth]{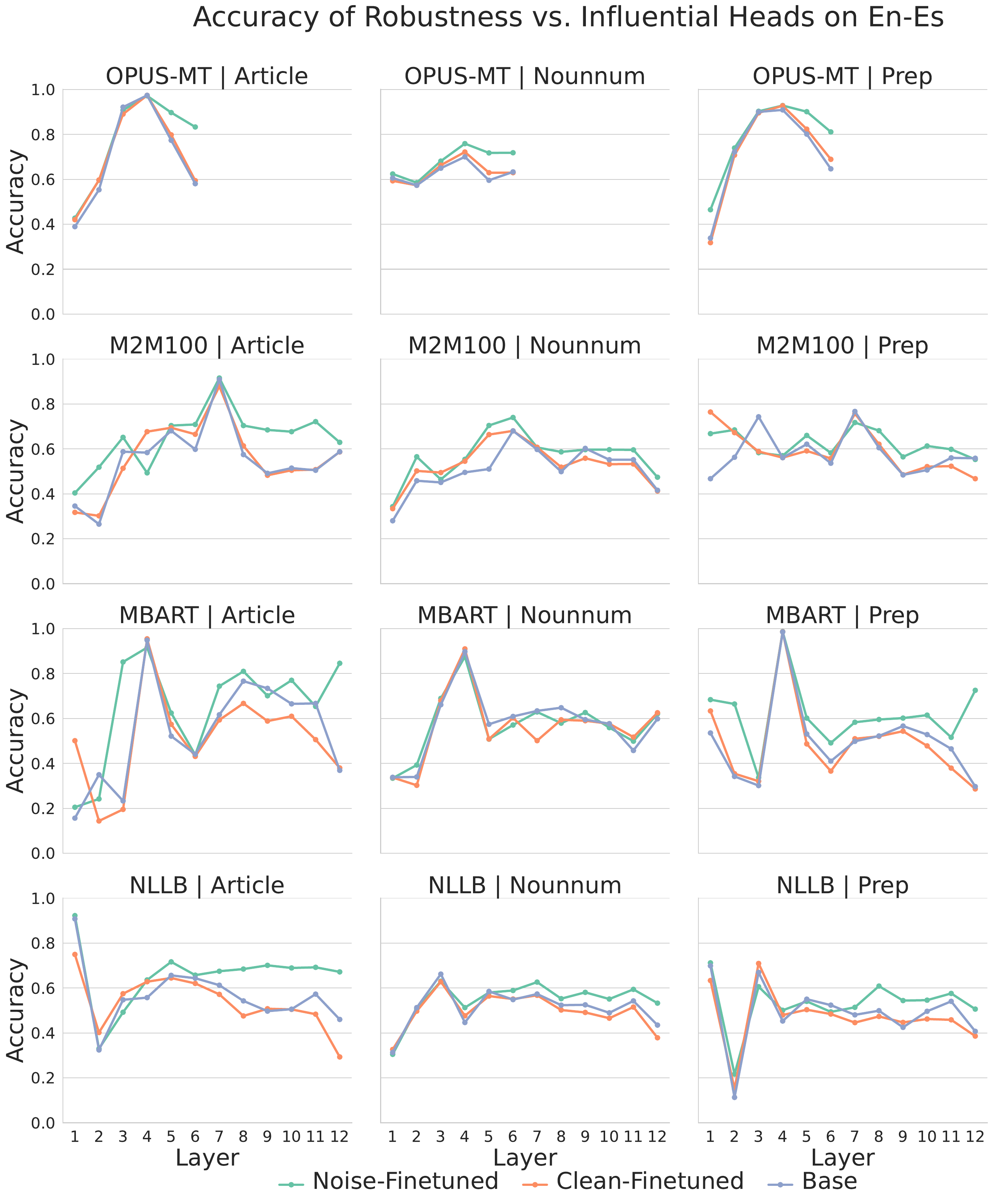}
    \caption{Accuracy of Robustness and Influential heads on En-Es. We find the accuracy is higher for Noise-Finetuned models especially in deep layers.}
    \label{fig:influential_robustness_heads}
\end{figure}
 
Base, Clean-Finetuned, and Noise-Finetuned models distribute their attention similarly to POS tags, but in some cases, the noisy model has learned to put more attention to the correct POS tags. For example, M2M100 model on Fr-Es has learned to put more attention on determiners and adjectives to deal with \textit{Nounnum} errors (going from a mean over layers of 0.021 and 0.030 to 0.023 and 0.035 respectively). On Fr-Es \textit{Article} errors, NLLB and OPUS-MT models learned to put more attention on nouns (from 0.051 and 0.065 to 0.053 and 0.069). On En-Es \textit{Article} errors, NLLB and OPUS-MT have learned to put more attention on nouns and proper nouns respectively (going from 0.052 and 0.057 to 0.055 and 0.061), while on \textit{Prep} errors, they have learned to put more attention on verbs (from 0.062 and 0.062 to 0.067 and 0.069). We see this trend in other language directions as well as across models and errors, although in some cases it is not very clear and the attention scores of \textit{Robustness Heads} of Noise-Finetuned and Base models are very close. 

\subsubsection{Similarity between Robustness and Influential Heads}
\label{robustness_influential_accuracy}

The previous section shows that Base, Clean-Finetuned and Noise-Finetuned models each contain \textit{Robustness Heads} that attend similarly to interpretable POS tags when dealing with grammatical errors. So what gives the Noise-Finetuned models an advantage in terms of robustness? To answer this question, we start by identifying \textit{Influential Heads} of the noisy word, as the heads that most influence it toward its current state, then we compare them with \textit{Robustness Heads}. Figure \ref{fig:influential_robustness_heads} presents the accuracy between \textit{Influential Heads} and \textit{Robustness Heads} at each layer of the encoder. The figure shows that this accuracy is higher in Noise-Finetuned models especially in deeper layers, which means that models after fine-tuning on noise, tend to employ more \textit{Robustness Heads} for updating the noisy word representation, and this can explain their improved robustness.

\section{Discussion}

\noindent
\textbf{RQ1: \textit{How do models represent and handle grammatical errors to achieve robustness?}} The encoder implements a GEC setup where it first detects the error then corrects it, and this behavior is more distinguishable in Noise-Finetuned models. Compared to Base or Clean-Finetuned models, Noise-Finetuned models maintain their error detection in lower layers, while they drive the representation of the ungrammatical word to be as closely similar to the grammatical form (\S\ref{representation_distance}). We argue that the correction component becomes more prominent later in the model, leading to lower GED probing accuracy and increased usage of \textit{Robustness Heads} in deeper layers (\S\ref{ged_probing} and \S\ref{robustness_influential_accuracy}). This finding coupled with the results in Appendix \ref{freezing} show that fine-tuning only the encoder while freezing the decoder is sufficient for achieving robustness, while also preserving performance on clean data.

\noindent
\textbf{RQ2: \textit{How does grammatical error representation differ across models and languages?}} 
\citet{davis-etal-2022-probing} suggest that GED probing performance can reflect the linguistic knowledge of pretrained models. While this might be true, our evidence indicates that it is more influenced by the linguistic features of the language itself (specifically the source language) (\S\ref{ged_probing}). For example, the GED probing performance of \textit{Nounnum} errors on Fr-Es peaks at around 0.8, while on En-Es, it peaks at maximum 0.5. This difference can be explained by the fact that in French articles and adjectives follow their noun in number, providing easy access to information about the noun in question. Furthermore, we find similarities across models in our analysis of representation distance and attention to POS tags as well (\S\ref{representation_distance} and \S\ref{attention_pos_tags}). However, when comparing across source languages, models represent and handle the same error differently. This suggests that languages might interfere with one another when fine-tuning for robustness across multiple languages, or even during training, which can be avoided by using language-specific adapters \cite{bapna-firat-2019-simple}. 

\noindent
\textbf{RQ3: \textit{How does fine-tuning for robustness influences models to achieve improved robustness?}} Fine-tuning for robustness leverages the existing knowledge of the Base models and generally does not go beyond it. If we look at probing performance, we see that generally, Noise-Finetuned models do not go beyond the peak in performance achieved by their Base model (\S\ref{ged_probing}), even though they are trained on the grammatical errors. Moreover, the \textit{Robustness Heads} of Noise-Finetuned models distribute their attention to POS tags similarly to their Base models (\S\ref{attention_pos_tags}). However, we find that Noise-Finetuned models rely more on \textit{Robustness Heads} especially in deeper layers to influence the noisy word toward the correct form (\S\ref{robustness_influential_accuracy}). This suggests that fine-tuning for robustness builds on the existing structure in pre-trained NMT models, therefore, analyzing the effects of different pre-training strategies and training data can be a valuable direction for future work.

\section{Conclusion and Future Work}
In this work, we analyze transformer NMT encoders under the effect of grammatical noise, and investigate how fine-tuning for robustness affects model behavior and internal representations. We find that the encoder -especially after fine-tuning- implements a GEC setup: it first detects the error and then corrects it by adjusting the representation of the ungrammatical word towards its correct form. To better understand this behavior, we propose a method for finding \textit{Robustness Heads}-attention heads that attend to POS tags and help detect and correct the grammatical error. Additionally, we find that fine-tuning on grammatical errors leads the model to use more \textit{Robustness Heads} especially in deeper layers.

These findings suggest a practical strategy for improving robustness in NMT systems: fine-tuning only the encoder on (synthetically) noisy data can substantially enhance robustness without degrading performance on clean data. This makes it an efficient and interpretable alternative to full model fine-tuning. In practice, such systems may benefit from selectively introducing common error types—especially those relevant to their deployment context—during fine-tuning. 

Furthermore, our analysis framework, which combines GED probing, representational similarity and influential attention heads, is model-agnostic and generalizable. It provides a systematic way to audit robustness across different models and languages and could be applied to other encoder-decoder NLP systems.

Although we focused our analysis on encoder-decoder models, we hypothesize that decoder-only models may exhibit a similar behavior especially in their early layers, therefore, we will extend this analysis to decoder only models in future work.

\section{Limitations}
Although Spanish, Italian and French belong to the Romance family, while English, German, and Dutch are Germanic languages. All of the languages belong to the Indo-European family, which might limit the generalization of our work to other languages. Additionally, we use synthetic noise which is more controllable and allows us to provide a fine-grained analysis of models, but there are definitely limitations in how close it can simulate natural noise. Finally, due to resource limitations, we focus on relatively small models, and while larger models might be more robust, we think that our analysis still offers interesting insights about robustness. 

\section*{Acknowledgments}
The research presented in this paper was conducted as part of VOXReality project\footnote{\texttt{\url{https://voxreality.eu/}}}, which was funded by the European Union Horizon Europe program under grant agreement No 101070521.

\bibliography{acl_latex}

\appendix

\section{Freezing Results}
\label{freezing}

\begin{figure*}[t]
    \centering
    \includegraphics[width=\textwidth]{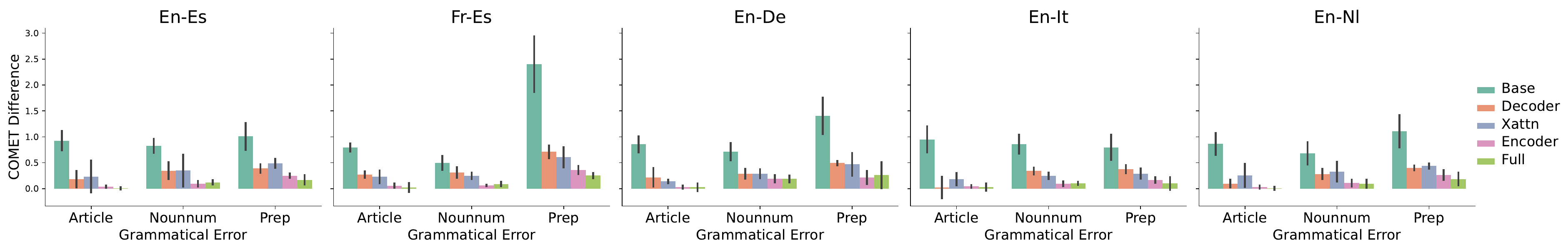}
    \caption{COMET difference between performance on clean and noisy test sets after fine-tuning different parts of the models. We show the average across 3 multilingual models (M2M100, MBART, NLLB) and bilingual models (OPUS-MT for each language pair). Fine-tuning the encoder is almost as good as fine-tuning the full model.}
    \label{fig:freezing_comet_scores}
\end{figure*}

When fine-tuning different parts of the model on grammatical errors, while keeping the other parts frozen, we find that fine-tuning only the encoder approaches fine-tuning the full model (both encoder and decoder) in terms of robustness, as shown in Figure \ref{fig:freezing_comet_scores}. Although the target translation is generated on the decoder side, fine-tuning the decoder does not lead to similar improvements as fine-tuning the encoder. This suggests that in NMT, robustness is more related to the source side representation than to the generation process, and therefore, we focus our analysis on the encoder.

\section{Experiments Details}

\subsection{Articles and Prepositions Lists}
\label{articles_prepositions}
We provide the list of articles and prepositions that we consider for our analysis:

\noindent
\textbf{English Articles:} \{a, an, the\}. \\
\textbf{English Prepositions:} \{on, in, at, from, for, under, over, with, into, during, until, against, among, throughout, of, to, by, about, like, before, after, since, across, behind, but, out, up, down, off\}.

\noindent
\textbf{French Articles:} \{la, le, un, une, les, des\}. \\
\textbf{French Prepositions:} \{à, après, avant, avec, chez, contre, dans, de, depuis, derrière, devant, durant, en, entre, envers, environ, jusque, malgré, par, parmi, pendant, pour, sans, sauf, selon, sous, suivant, sur, vers\}.

\subsection{Data Splits}
Table \ref{tab:dataset_splits} shows the amount of data for each language direction in our experiments. Although the train set for Fr-Es is small in comparison to the other language directions, the COMET improvements after fine-tuning shown in Table \ref{finetuning_results_appendix} suggest that it is enough for fine-tuning and potentially for drawing conclusions about its effects. The test sets which we use for evaluating models and for our analysis are almost similar in size.
\label{data_splits_appendix}
\begin{table}[h!]
    \centering
    \begin{tabular}{c|c|c|c}
        \toprule
        & \textbf{Train} & \textbf{Dev} & \textbf{Test} \\
        \midrule
        En-Es & 31607 & 1272 & 1267 \\
        En-De & 32628 & 1320 & 1253 \\
        En-Nl & 31401 & 1269 & 1235 \\
        En-It & 29552 & 1122 & 1130 \\
        Fr-Es & 7857  & 1072 & 1098 \\
        \bottomrule
    \end{tabular}
    \caption{Dataset splits for each language direction}
    \label{tab:dataset_splits}
\end{table}

\subsection{POS Tags List}
\label{pos_tags_list}
\textbf{ADJ:} Adjective. \\
\noindent
\textbf{ADP:} Adposition. \\
\noindent
\textbf{ADV:} Adverb. \\
\noindent
\textbf{AUX:} Auxiliary. \\
\noindent
\textbf{CCONJ:} Coordinating conjunction. \\
\noindent
\textbf{DET:} Determiner. \\
\noindent
\textbf{NOUN:} Noun. \\
\noindent
\textbf{PRON:} Pronoun. \\
\noindent
\textbf{PROPN:} Proper noun. \\
\noindent
\textbf{PUNCT:} Punctuation. \\
\noindent
\textbf{VERB:} Verb. \\

\section{Additional Results}

\subsection{Fine-tuning Results}
\label{finetuning_results_appendix}
We present the results of fine-tuning NMT models for robustness on Fr-Es, En-De, En-It and En-Nl in Table \ref{table:finetuning_results_appendix}.

\begin{table*}[tbh]
\scriptsize
\centering
\begin{tabular}{llrrrrrrrrrr}
\midrule
& & \multicolumn{3}{c}{\textbf{Article}} & \multicolumn{3}{c}{\textbf{Nounnum}} & \multicolumn{3}{c}{\textbf{Prep}} \\
\textbf{Direction} & \textbf{Model} & \textbf{Clean} & \textbf{Noisy} & \textbf{$\Delta$} & \textbf{Clean} & \textbf{Noisy} & \textbf{$\Delta$} & \textbf{Clean} & \textbf{Noisy} & \textbf{$\Delta$} \\
\midrule
\multirow{13}{*}{Fr-Es}  
& opus-mt-base  & 74.57   & 73.79   & 0.79       & 74.57   & 73.98   & 0.6        & 74.57 & 71.88 & 2.7        \\
& opus-mt-clean & 75.36   & 74.7    & 0.66       & 75.13   & 74.7    & 0.43       & \textbf{75.7}  & 73.11 & 2.58       \\
& opus-mt-noise & \textbf{75.62}   & \textbf{75.54}   & \textbf{0.07}       & \textbf{75.69}   & \textbf{75.62}   & \textbf{0.07}       & 75.65 & \textbf{75.18} & \textbf{0.46} \\
\cmidrule{2-11}
& m2m100-base   & 73.04   & 72.21   & 0.83       & 73.04   & 72.45   & 0.59       & 73.04 & 70.06 & 2.98       \\
& m2m100-clean  & 74.01   & 73.45   & 0.55       & 73.98   & 73.65   & 0.33       & \textbf{74.27} & 72.21 & 2.07       \\
& m2m100-noise  & \textbf{74.24}   & \textbf{74.23}   & \textbf{0.01}       & \textbf{74.22}   & \textbf{74.16}   & \textbf{0.06}       & 74.2  & \textbf{73.84} & \textbf{0.36}       \\
\cmidrule{2-11}
& mbart-base    & 69.13   & 68.27   & 0.87       & 69.13   & 68.82   & 0.31       & 69.13 & 67.02 & 2.12       \\
& mbart-clean   & \textbf{74.25}   & \textbf{73.72}   & 0.52       & 74.28   & 73.7    & 0.58       & 74.17 & 72.41 & 1.76       \\
& mbart-noise   & 73.68   & 73.57   & \textbf{0.11}       & \textbf{74.33}   & \textbf{74.29}   & \textbf{0.04}       & \textbf{74.23} & \textbf{73.9}  & \textbf{0.33}       \\
\cmidrule{2-11}
& nllb-base     & 74.92   & 74.23   & 0.69       & 74.92   & 74.44   & 0.48       & 74.92 & 73.12 & 1.8        \\
& nllb-clean    & \textbf{76.1}    & 75.55   & 0.55       & 76.03   & 75.62   & 0.41       & 75.99 & 74.68 & 1.31       \\
& nllb-noise    & 76.0    & \textbf{75.97}   & \textbf{0.04}       & \textbf{76.12}   & \textbf{76.04}   & \textbf{0.08}       & \textbf{76.07} & \textbf{75.79} & \textbf{0.28}       \\
\midrule
\multirow{13}{*}{En-De}     
& opus-mt-base  & 76.93   & 75.89   & 1.04       & 76.93   & 76.0    & 0.93       & 76.93 & 75.22 & 1.71       \\
& opus-mt-clean & 77.83   & 76.91   & 0.93       & 77.82   & 77.03   & 0.79       & 77.89 & 76.47 & 1.42       \\
& opus-mt-noise & \textbf{77.86}   & \textbf{77.86}   & \textbf{0.0}       & \textbf{77.93}   & \textbf{77.73}   & \textbf{0.2}        & \textbf{77.93} & \textbf{77.54} & \textbf{0.4}        \\
\cmidrule{2-11}
& m2m100-base   & 72.94   & 72.05   & 0.89       & 72.94   & 72.19   & 0.75       & 72.94 & 71.42 & 1.52       \\
& m2m100-clean  & 75.29   & 74.64   & 0.65       & 75.08   & 74.31   & 0.78       & 75.1  & 74.02 & 1.09       \\
& m2m100-noise  & \textbf{75.61}   & \textbf{75.54}   & \textbf{0.07}       & \textbf{75.42}   & \textbf{75.14}   & \textbf{0.28}       & \textbf{75.31} & \textbf{75.14} & \textbf{0.17}       \\
\cmidrule{2-11}
& mbart-base    & 76.26   & 75.44   & 0.82       & 76.26   & 75.64   & 0.62       & 76.26 & 74.78 & 1.48       \\
& mbart-clean   & 77.29   & 76.84   & 0.45       & 77.35   & 76.87   & 0.48       & \textbf{77.54} & 76.66 & 0.88       \\
& mbart-noise   & \textbf{77.5}    & \textbf{77.49}   & \textbf{0.02}       & \textbf{77.5}    & \textbf{77.35}   & \textbf{0.15}       & 77.47 & \textbf{77.34} & \textbf{0.13}       \\
\cmidrule{2-11}
& nllb-base     & 76.7    & 76.03   & 0.67       & 76.7    & 76.15   & 0.55       & 76.7  & 75.79 & 0.9        \\
& nllb-clean    & 77.52   & 76.96   & 0.56       & 77.49   & 77.05   & 0.43       & 77.6  & 76.72 & 0.88       \\
& nllb-noise    & \textbf{77.56}   & \textbf{77.52}   & \textbf{0.04}       & \textbf{77.51}   & \textbf{77.39}   & \textbf{0.12}       & \textbf{77.61} & \textbf{77.45} & \textbf{0.16}       \\
\midrule
\multirow{13}{*}{En-It}       
& opus-mt-base  & 77.22   & 76.08   & 1.14       & 77.22   & 76.26   & 0.95       & 77.22 & 76.13 & 1.09       \\
& opus-mt-clean & 77.31   & 76.35   & 0.96       & 77.45   & 76.53   & 0.92       & 77.42 & 76.37 & 1.05       \\
& opus-mt-noise & \textbf{77.53}   & \textbf{77.46}   & \textbf{0.07}       & \textbf{77.71}   & \textbf{77.58}   & \textbf{0.13}       & \textbf{77.7}  & \textbf{77.51} & \textbf{0.2}        \\
\cmidrule{2-11}
& m2m100-base   & 75.0    & 73.82   & 1.19       & 75.0    & 73.95   & 1.06       & 75.0  & 74.11 & 0.89       \\
& m2m100-clean  & \textbf{76.83}   & 75.77   & 1.06       & 76.5    & 75.51   & 0.99       & 76.61 & 75.88 & 0.73       \\
& m2m100-noise  & 76.81   & \textbf{76.8}    & \textbf{0.01}       & \textbf{76.83}   & \textbf{76.8}    & \textbf{0.03}       & \textbf{76.91} & \textbf{76.75} & \textbf{0.16}       \\
\cmidrule{2-11}
& mbart-base    & 75.73   & 74.97   & 0.75       & 75.73   & 75.08   & 0.65       & 75.73 & 75.04 & 0.68       \\
& mbart-clean   & \textbf{77.63}   & 76.88   & 0.76       & \textbf{77.75}   & 77.01   & 0.74       & \textbf{77.67} & 77.18 & 0.48       \\
& mbart-noise   & 77.58   & \textbf{77.53}   & \textbf{0.05}       & 77.73   & \textbf{77.62}   & \textbf{0.11}       & 77.62 & \textbf{77.41} & \textbf{0.21}       \\
\cmidrule{2-11}
& nllb-base     & 77.47   & 76.75   & 0.72       & 77.47   & 76.7    & 0.77       & 77.47 & 76.93 & 0.53       \\
& nllb-clean    & \textbf{78.07}   & 77.45   & 0.61       & \textbf{77.94}   & 77.35   & 0.59       & \textbf{77.94} & 77.53 & 0.41       \\
& nllb-noise    & 77.96   & \textbf{77.92}   & \textbf{0.04}       & 77.88   & \textbf{77.77}   & \textbf{0.1}        & 77.86 & \textbf{77.76} & \textbf{0.09}       \\
\midrule
\multirow{13}{*}{En-Nl}       
& opus-mt-base  & 78.03   & 76.99   & 1.03       & 78.03   & 77.13   & 0.89       & 78.03 & 76.62 & 1.41       \\
& opus-mt-clean & 78.66   & 77.76   & 0.9        & 78.65   & 77.83   & 0.82       & 78.86 & 77.61 & 1.25       \\
& opus-mt-noise & \textbf{79.05}   & \textbf{79.0}    & \textbf{0.06}       & \textbf{78.93}   & \textbf{78.8}    & \textbf{0.14}       & \textbf{78.98} & \textbf{78.62} & \textbf{0.36}  \\
\cmidrule{2-11}
& m2m100-base   & 74.9    & 73.91   & 0.98       & 74.9    & 74.08   & 0.82       & 74.9  & 73.59 & 1.31       \\
& m2m100-clean  & 76.67   & 75.82   & 0.86       & 76.75   & 75.89   & 0.86       & 76.85 & 75.93 & 0.92       \\
& m2m100-noise  & \textbf{76.78}   & \textbf{76.79}   & \textbf{-0.01}      & \textbf{76.9}    & \textbf{76.81}   & \textbf{0.08}       & \textbf{76.99} & \textbf{76.66} & \textbf{0.32}       \\
\cmidrule{2-11}
& mbart-base    & 74.86   & 73.97   & 0.89       & 74.86   & 74.26   & 0.6        & 74.86 & 73.89 & 0.97       \\
& mbart-clean   & \textbf{77.82}   & 77.12   & 0.71       & 77.86   & 77.08   & 0.79       & 78.05 & 77.22 & 0.83       \\
& mbart-noise   & 77.77   & \textbf{77.73}   & \textbf{0.04}       & \textbf{77.98}   & \textbf{77.8}    & \textbf{0.18}       & \textbf{78.08} & \textbf{77.93} & \textbf{0.15}       \\
\cmidrule{2-11}
& nllb-base     & 77.95   & 77.4    & 0.56       & 77.95   & 77.53   & 0.42       & 77.95 & 77.21 & 0.74       \\
& nllb-clean    & \textbf{78.85}   & 78.26   & 0.59       & \textbf{78.81}   & 78.3    & 0.45       & \textbf{78.74} & 78.12 & 0.63       \\
& nllb-noise    & 78.74   & \textbf{78.72}   & \textbf{0.02}       & 78.7    & \textbf{78.67}   & \textbf{0.03}       & 78.73 & \textbf{78.52} & \textbf{0.22}       \\
\bottomrule
\end{tabular}
\caption{COMET scores on Fr-Es, En-De, En-It and En-NL. The Base model is the original model, Clean is fine-tuned on the clean version of the data, and Noise is fine-tuned on the noisy version (with the same noise as the one they are tested on). We present the performance on the clean and noisy test sets and their difference ($\Delta$).}
\label{table:finetuning_results_appendix}
\end{table*}

\subsection{GED Probing}
\label{ged_probing_appendix}
\begin{figure*}[htbp]
    \centering
    \subfigure[Fr-Es]{
        \includegraphics[width=0.48\textwidth]{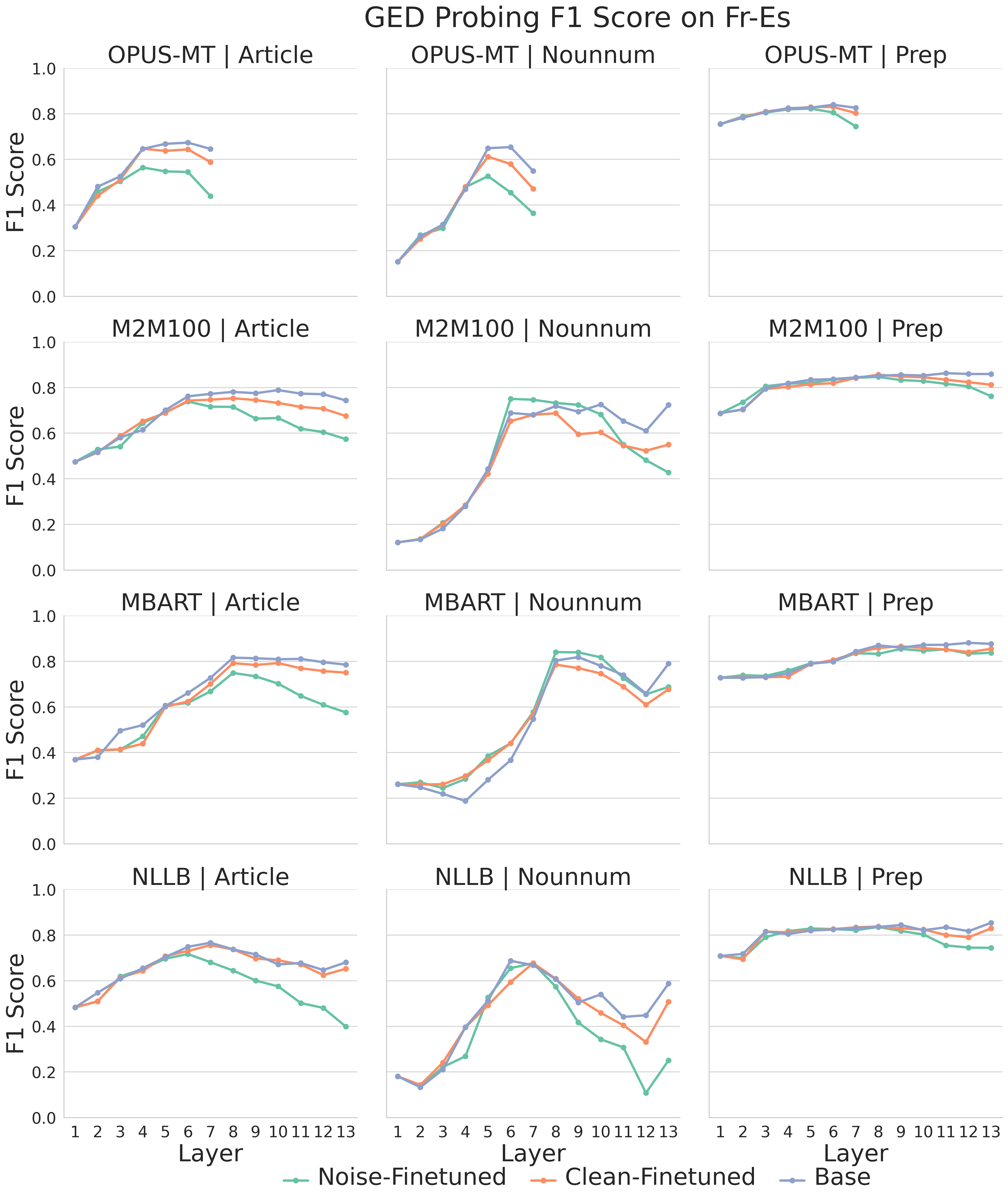}
        \label{fig:ged_probing_fr_es}
    }
    \subfigure[En-De]{
        \includegraphics[width=0.48\textwidth]{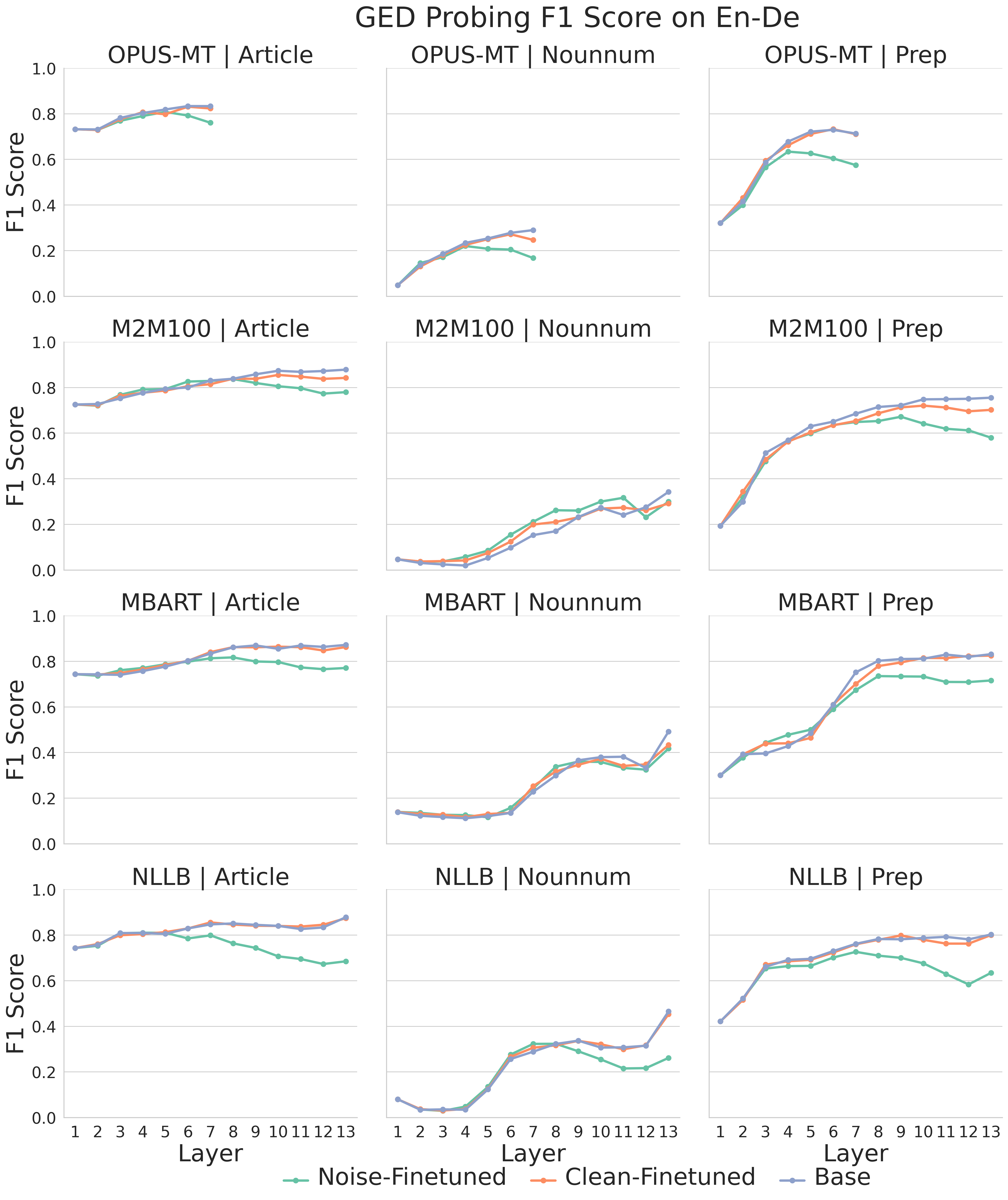}
        \label{fig:ged_probing_en_de}
    }
    \subfigure[En-It]{
        \includegraphics[width=0.48\textwidth]{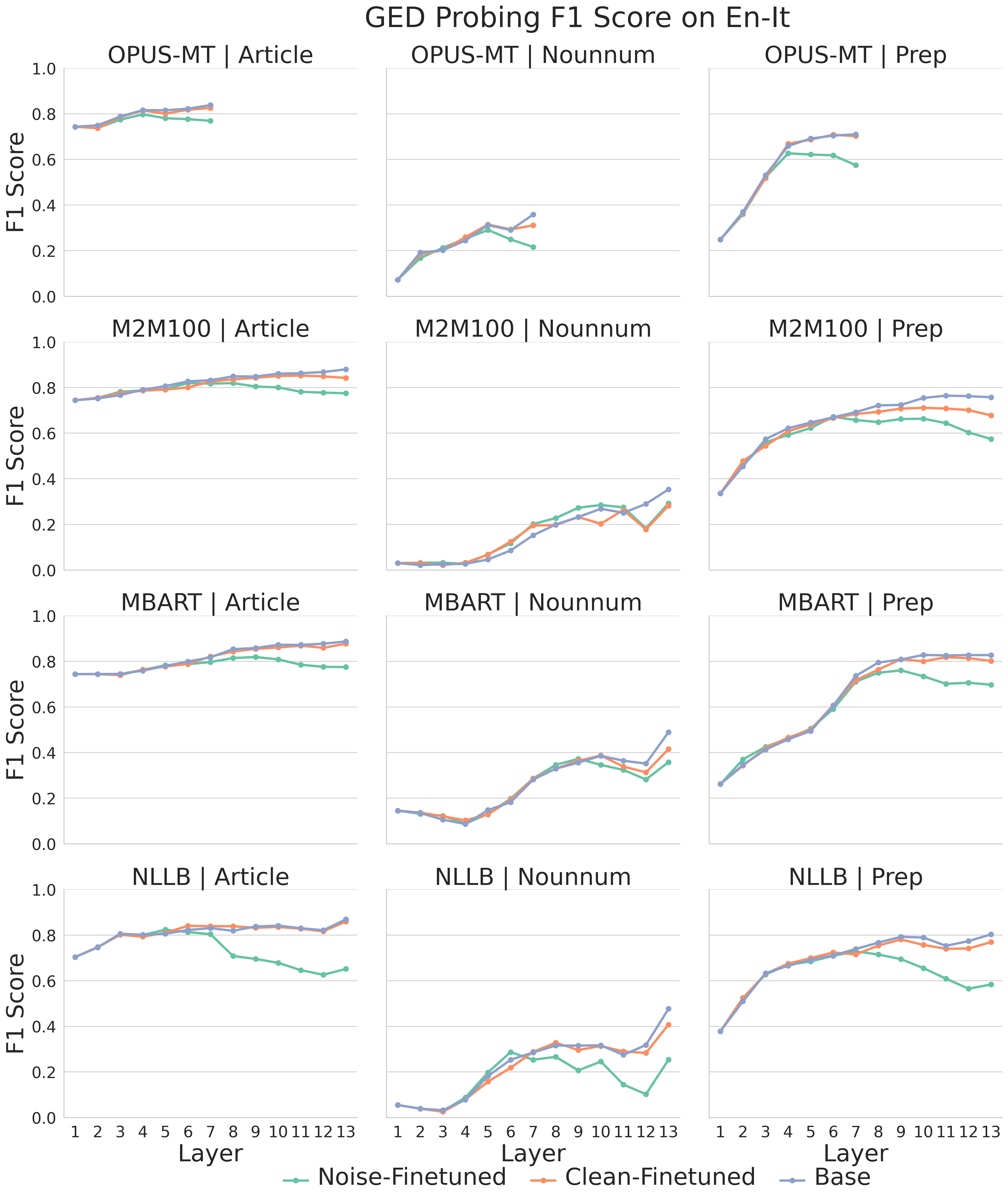}
        \label{fig:ged_probing_en_it}
    }
    \subfigure[En-Nl]{
        \includegraphics[width=0.48\textwidth]{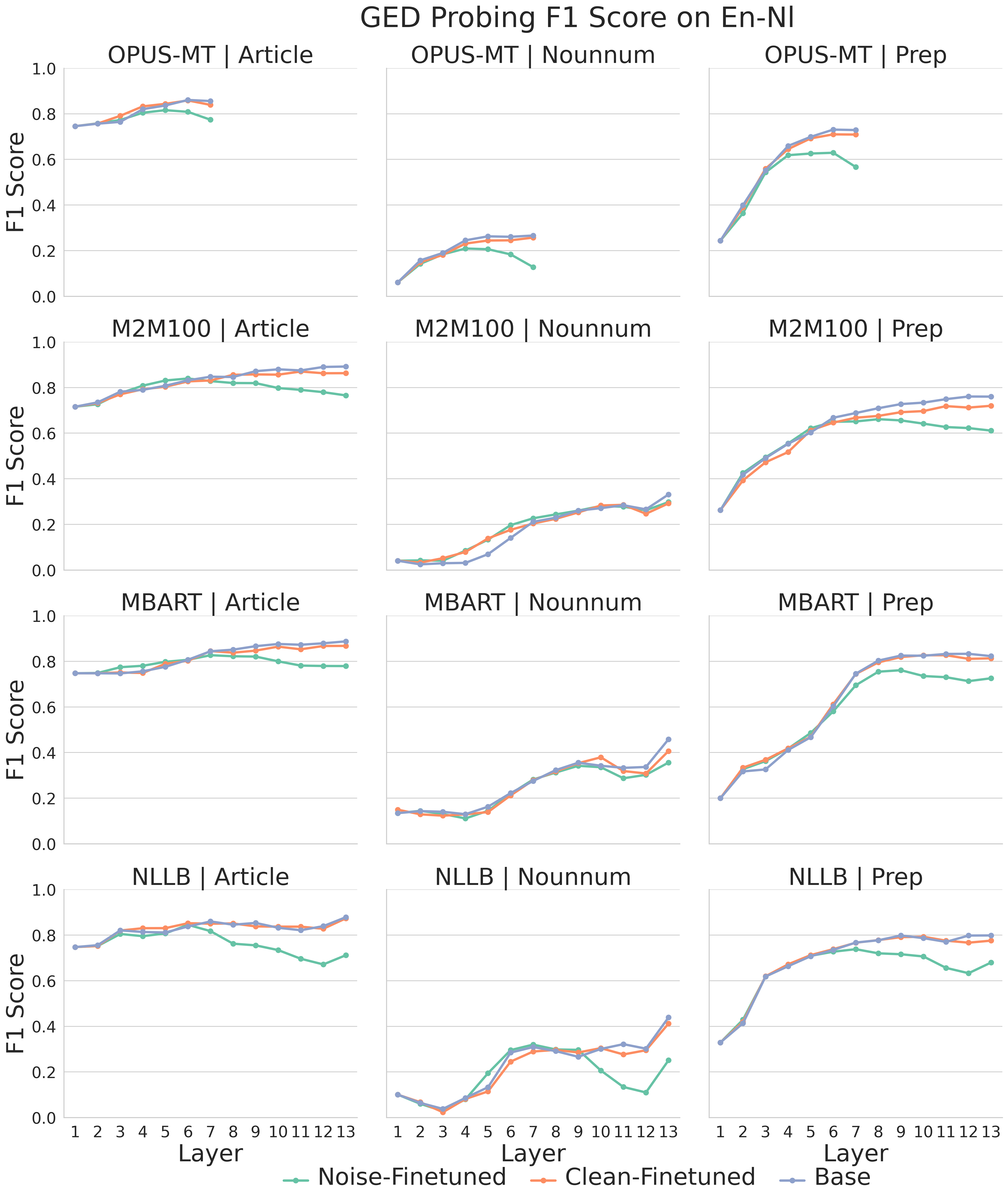}
        \label{fig:ged_probing_en_nl}
    }
    \caption{GED probing performance of Noise-Finetuned, Clean-Finetuned and Base models on Fr-Es, En-De, En-It and En-Nl. GED probing performance of Noise-Finetuned models witnesses a degradation in deeper layers.}
    \label{fig:ged_probing_appendix}
\end{figure*}

Figure \ref{fig:ged_probing_appendix} shows the GED probing performance on Fr-Es, En-De, En-It, and En-Nl.

\subsection{Respresentation Similarity}
\label{cka_distance_appendix}
\begin{figure*}[htbp]
\centering
    \subfigure[Fr-Es]{
        \includegraphics[width=0.48\textwidth]{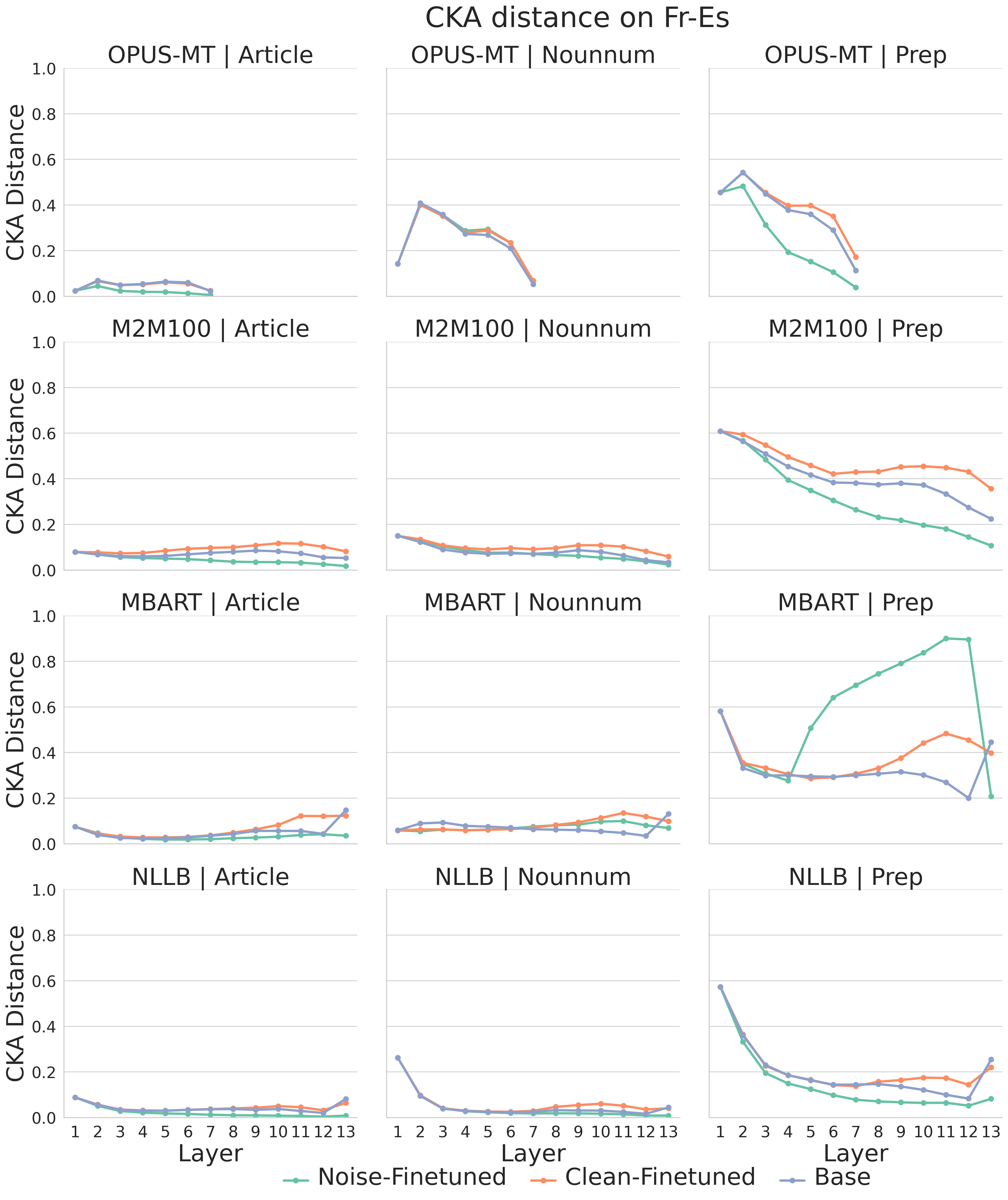}
        \label{fig:cka_distance_fr_es}
        }
        \centering
        \subfigure[En-De]{
        \includegraphics[width=0.48\textwidth]{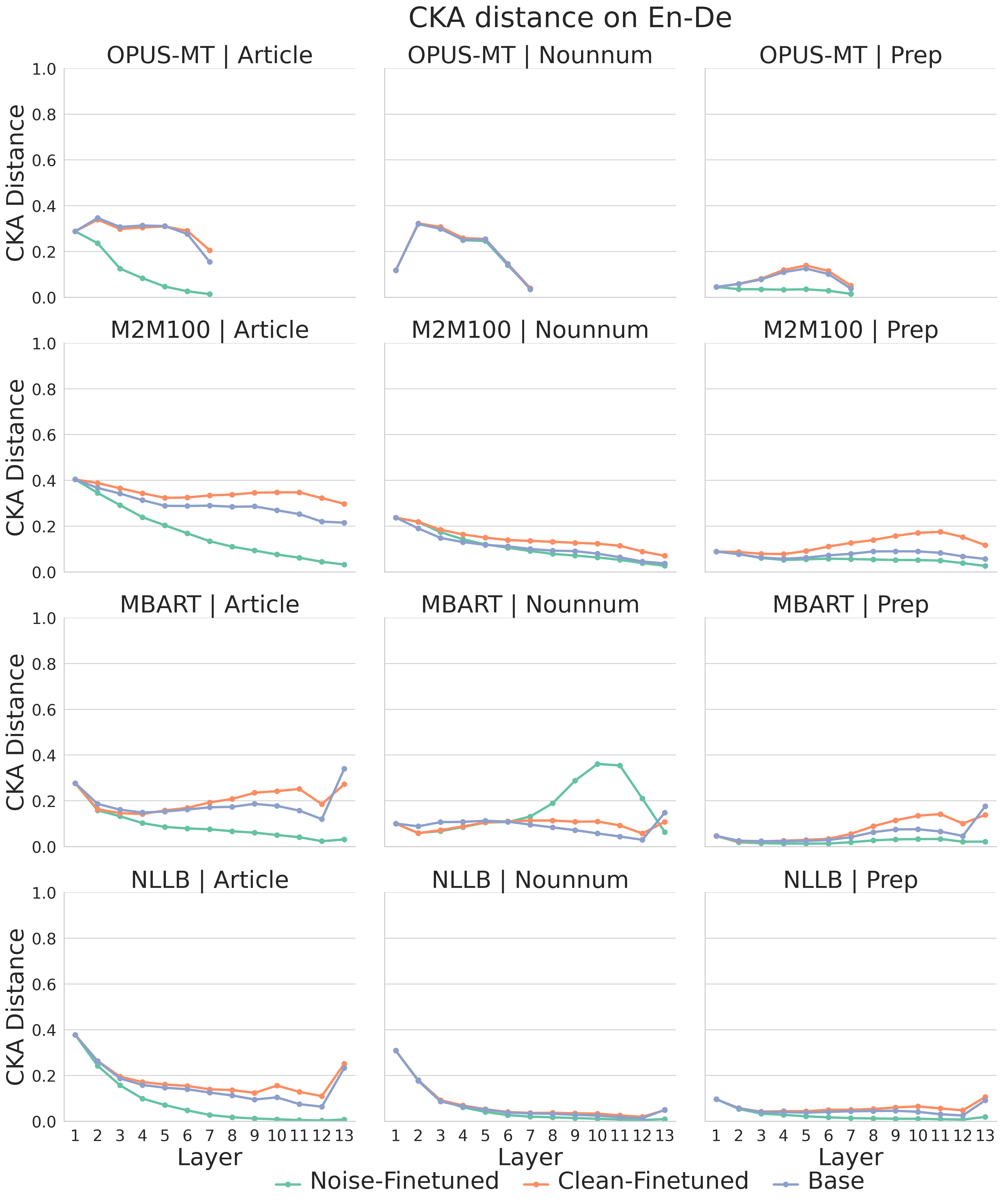}
        \label{fig:cka_distance_en_de}
        }
        \centering
        \subfigure[En-It]{
        \includegraphics[width=0.48\textwidth]{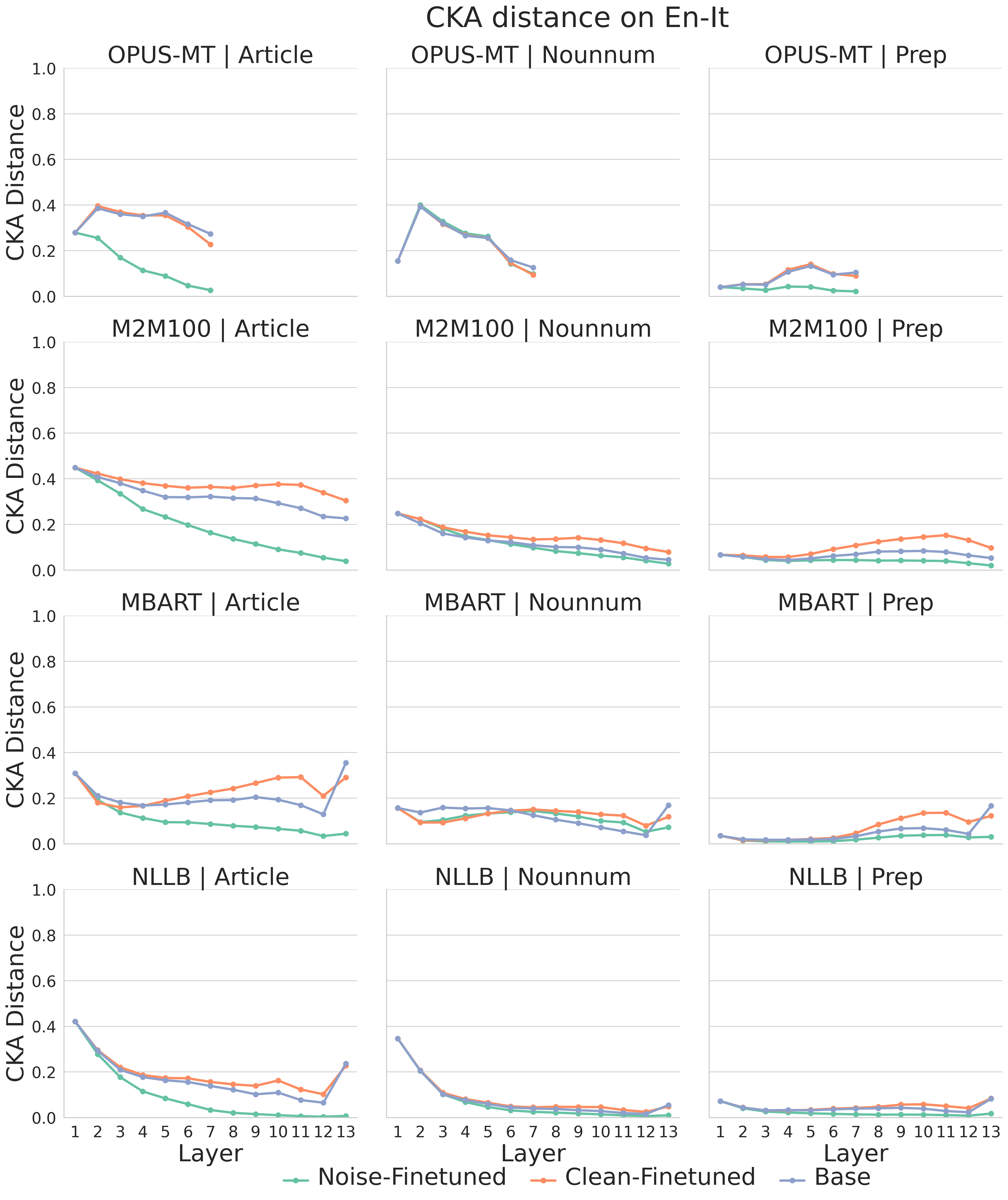}
        \label{fig:cka_distance_en_it}
        }
        \centering
        \subfigure[En-Nl]{
        \includegraphics[width=0.48\textwidth]{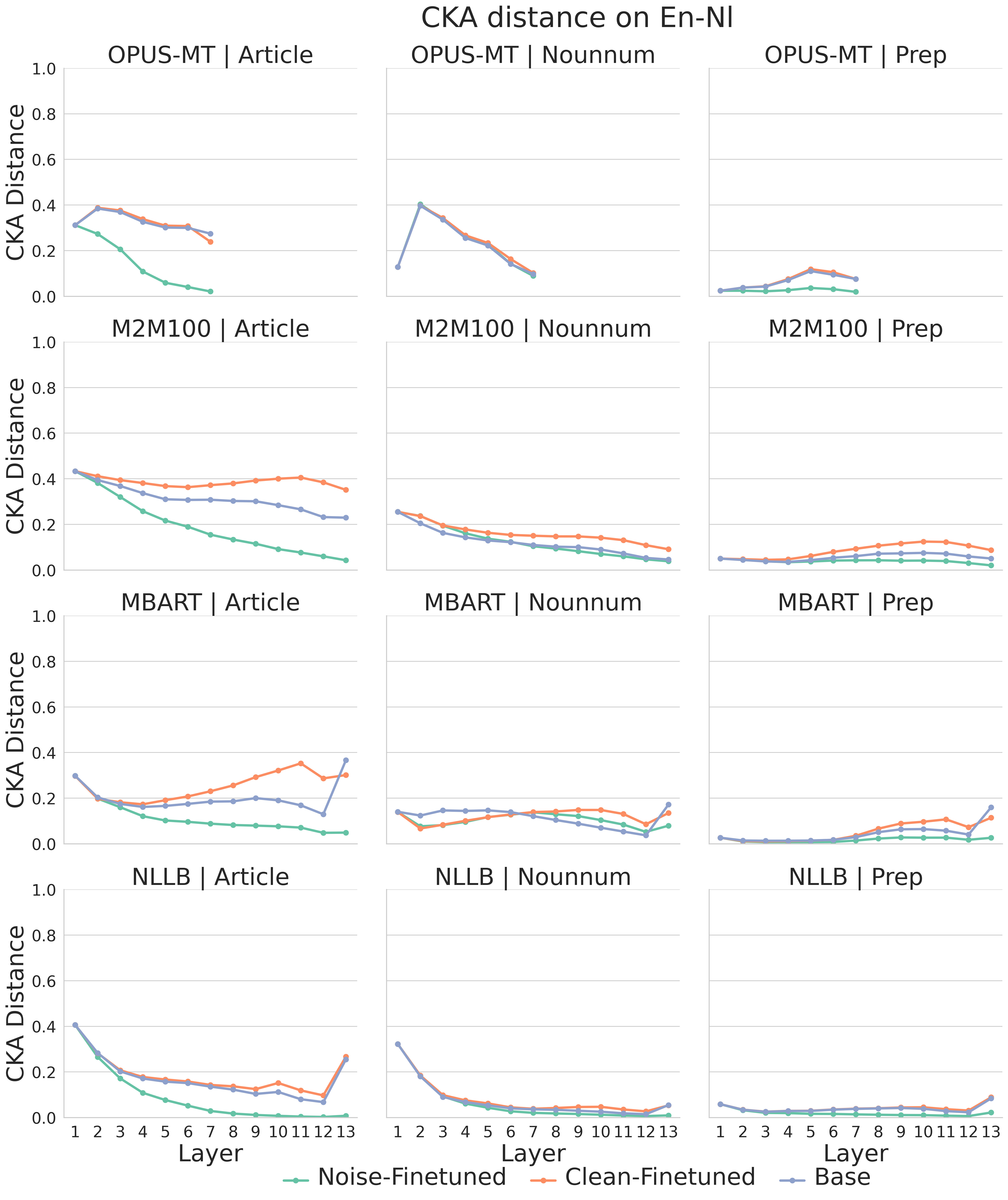}
        \label{fig:cka_distance_en_nl}
        }
    \caption{CKA distance of clean and noise word representations across models and errors on Fr-Es, En-De, En-It and En-Nl. Noise-Finetuned models drive the representations of the noisy word to be more similar to the clean word.}
    \label{fig:cka_distance_appendix}
\end{figure*}

Figures \ref{fig:cka_distance_appendix} shows the CKA distance of clean and noisy word representations on Fr-Es, En-De, En-It and En-Nl.

\subsection{Attention to POS Tags}
\label{attention_pos_appendix}
\begin{figure*}[t]
    \centering
    \includegraphics[width=0.95\textwidth,height=0.42\textheight]{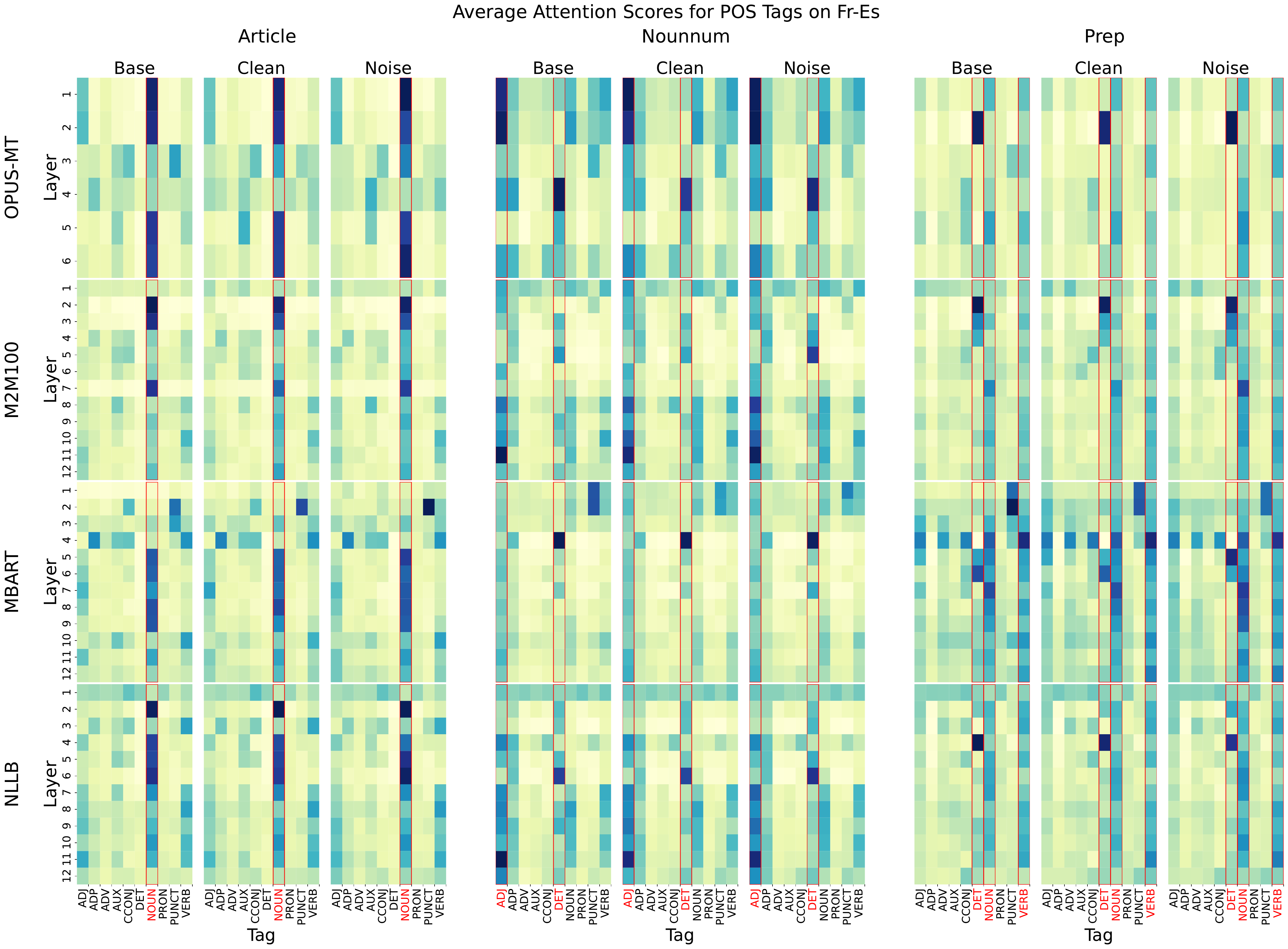}
    \caption{\textit{Robustness Heads} attention to the 10 most common POS tags in the test set on Fr-Es. The scale of attention is relative to each base model and error. We highlight POS tags that are attended to the most across models.}
    \label{fig:attention_pos_fr_es}
\vspace{-0.1in}  
\end{figure*}

\begin{figure*}[t]
    \centering
    \includegraphics[width=0.95\textwidth,height=0.42\textheight]{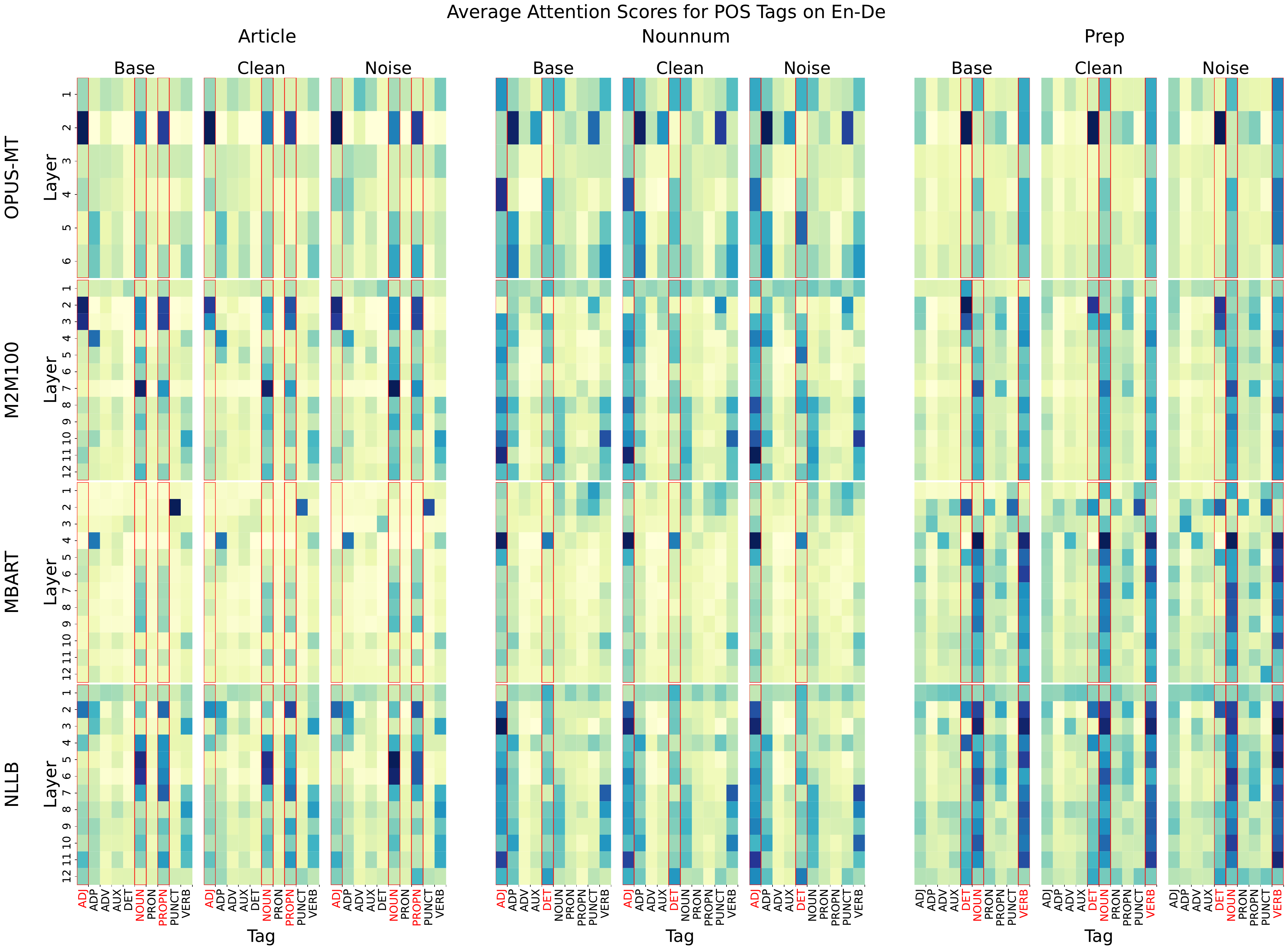}
    \caption{\textit{Robustness Heads} attention to the 10 most common POS tags in the test set on En-De. The scale of attention is relative to each base model and error. We highlight POS tags that are attended to the most across models.}
    \label{fig:attention_pos_en_de}
\end{figure*}

\begin{figure*}[t]
    \centering
    \includegraphics[width=0.95\textwidth,height=0.42\textheight]{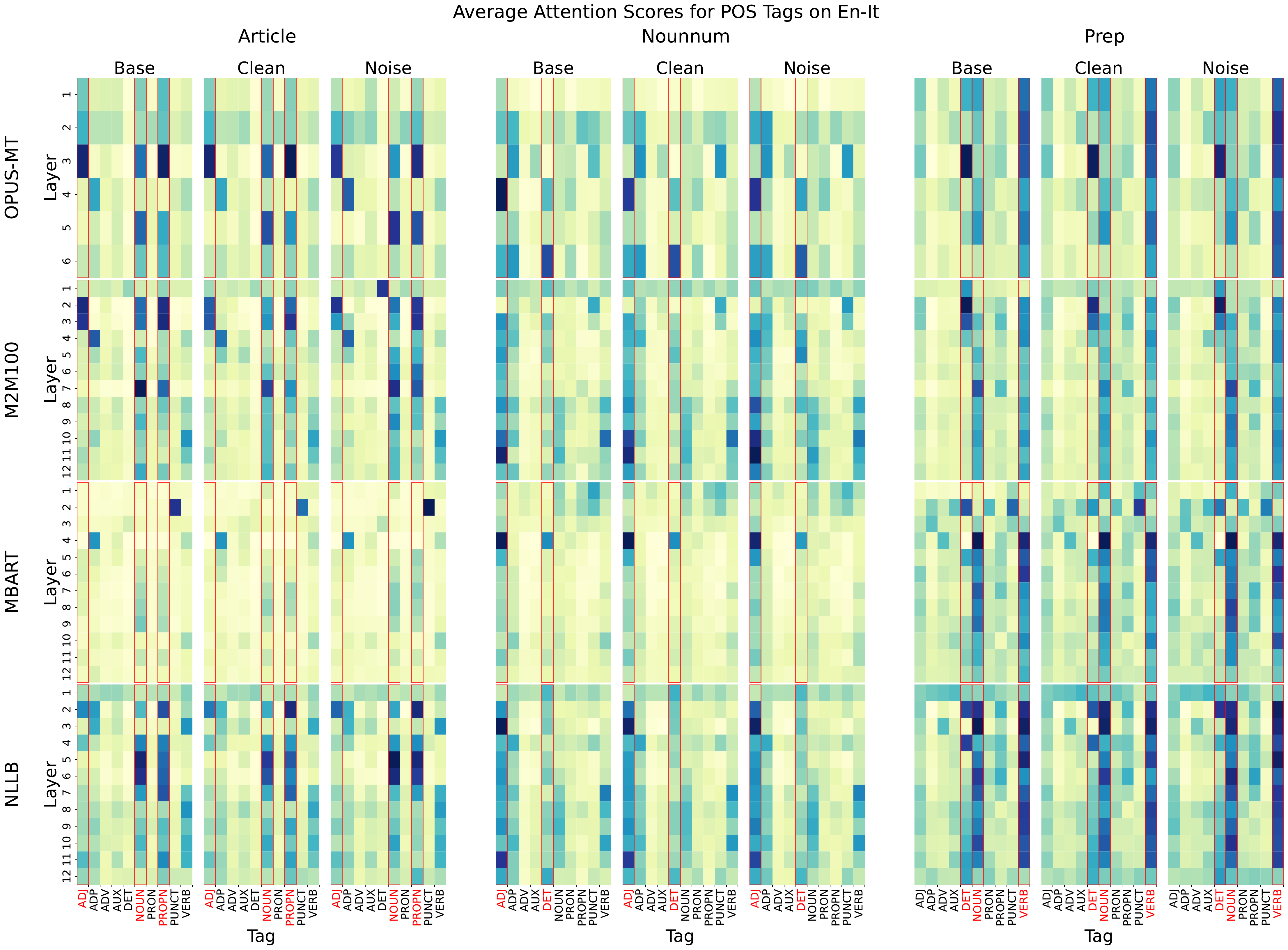}
    \caption{\textit{Robustness Heads} attention to the 10 most common POS tags in the test set on En-It. The scale of attention is relative to each base model and error. We highlight POS tags that are attended to the most across models.}
    \label{fig:attention_pos_en_it}
\vspace{-0.1in}  
\end{figure*}

\begin{figure*}[t]
    \centering
    \includegraphics[width=0.95\textwidth,height=0.42\textheight]{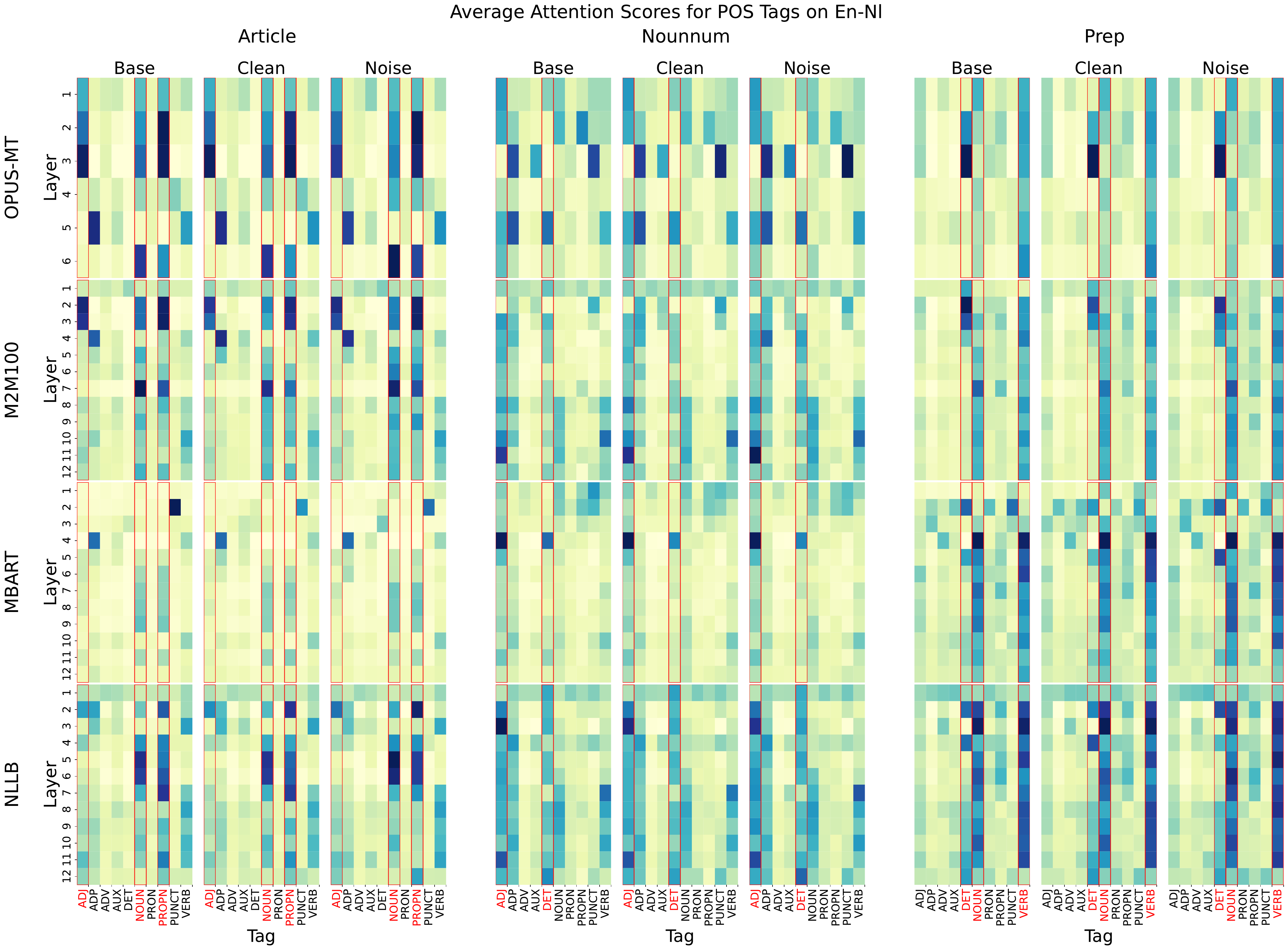}
    \caption{\textit{Robustness Heads} attention to the 10 most common POS tags in the test set on En-Nl. The scale of attention is relative to each base model and error. We highlight POS tags that are attended to the most across models.}
    \label{fig:attention_pos_en_nl}
\end{figure*}

Figure \label{} shows the accuracy of Robustness and Influential heads on \textit{Morpheus} errors.
Figures \ref{fig:attention_pos_fr_es}, \ref{fig:attention_pos_en_de}, \ref{fig:attention_pos_en_it} and \ref{fig:attention_pos_en_nl} show \textit{Robustness Heads} attention to POS tags on Fr-Es, En-De, En-It and En-Nl respectively.

\subsection{Similarity between Robustness and Influential Heads}
\label{robustness_influential_accuracy_appendix}
\begin{figure*}[ht]
    \centering
    \subfigure[Fr-Es]{%
        \includegraphics[width=0.48\textwidth]{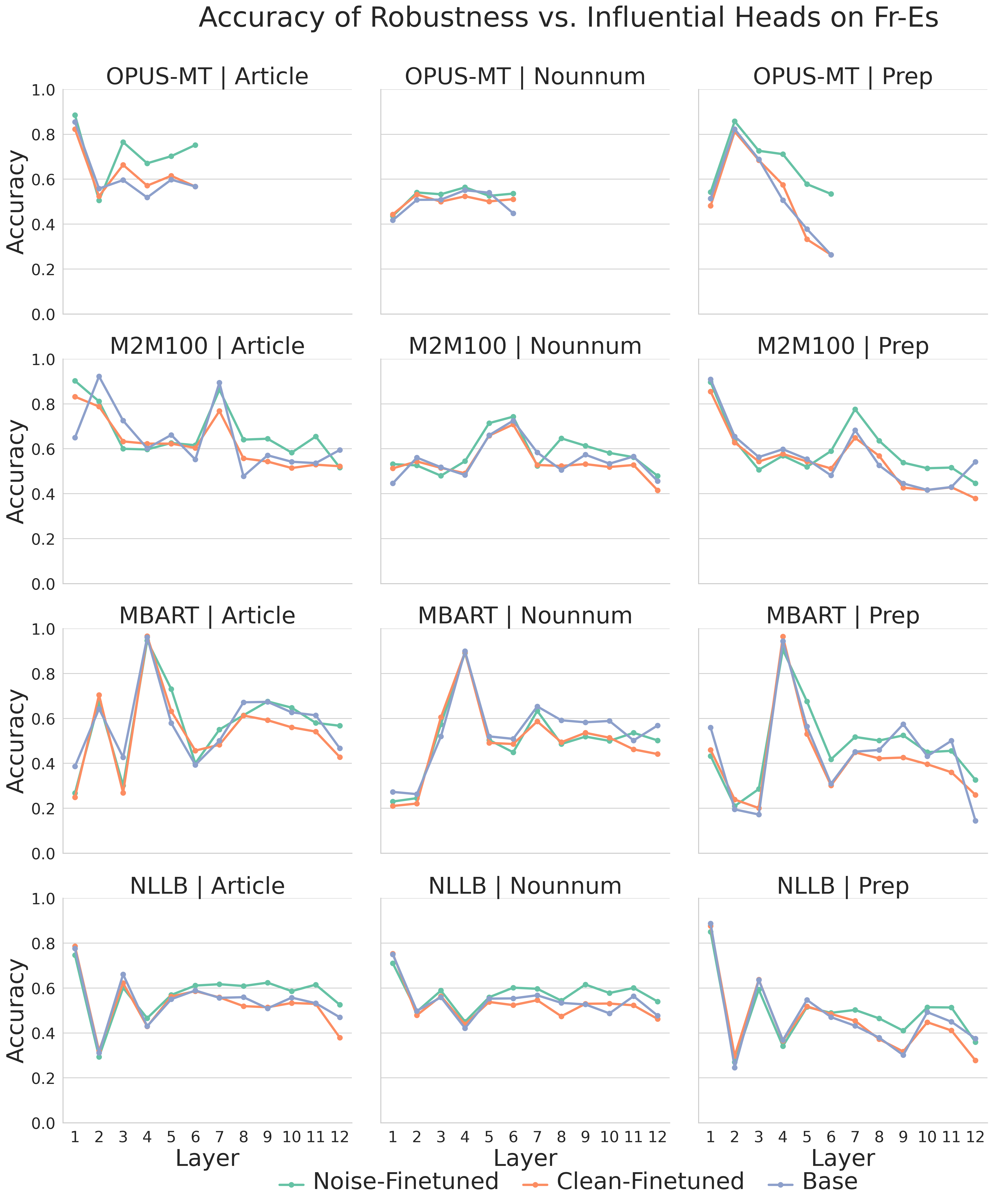}
        \label{fig:influential_robustness_heads_fr_es}
    }
    \hfill
    \subfigure[En-De]{%
        \includegraphics[width=0.48\textwidth]{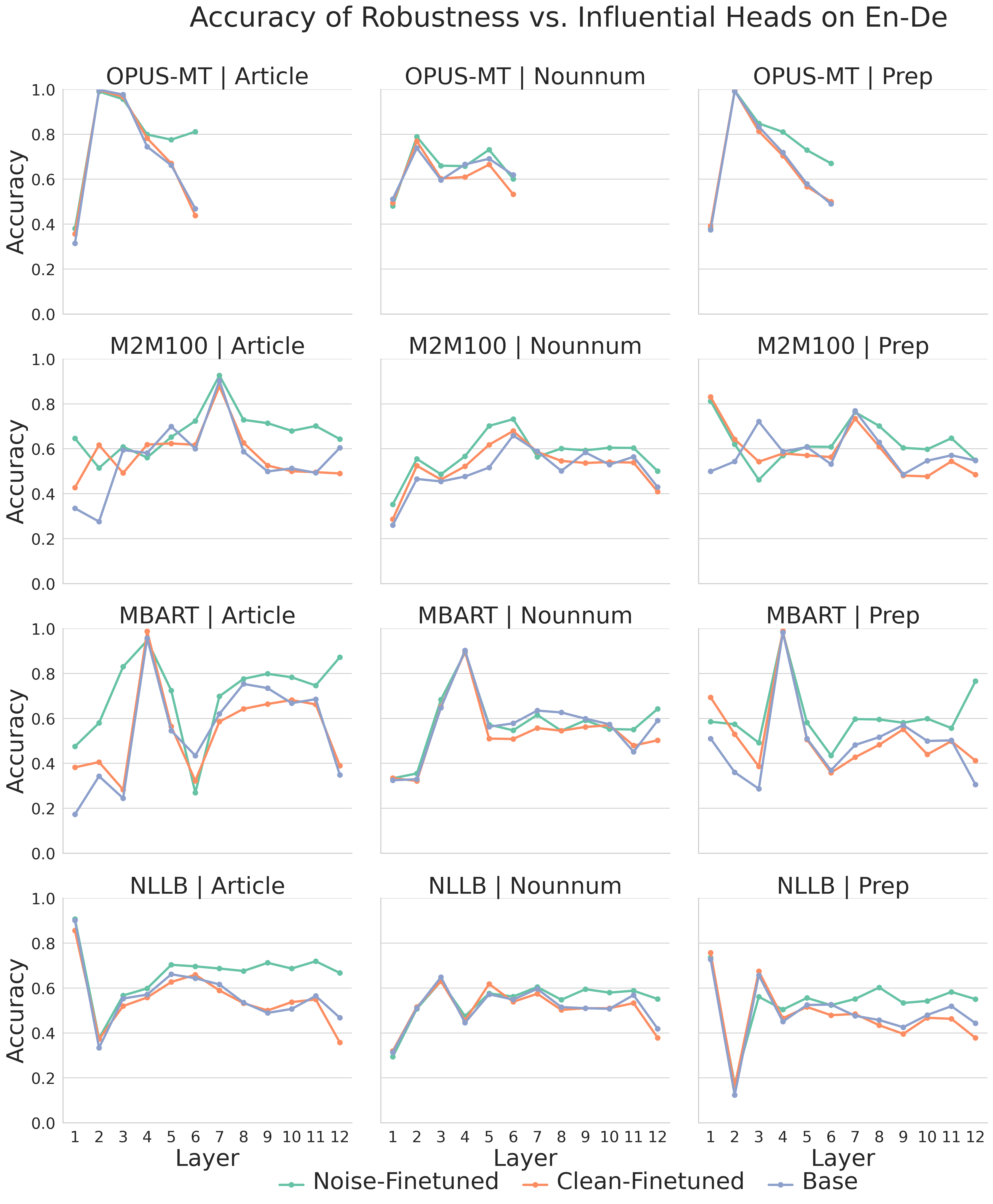}
        \label{fig:influential_robustness_heads_en_de}
    }
    \vskip\baselineskip
    \subfigure[En-It]{%
        \includegraphics[width=0.48\textwidth]{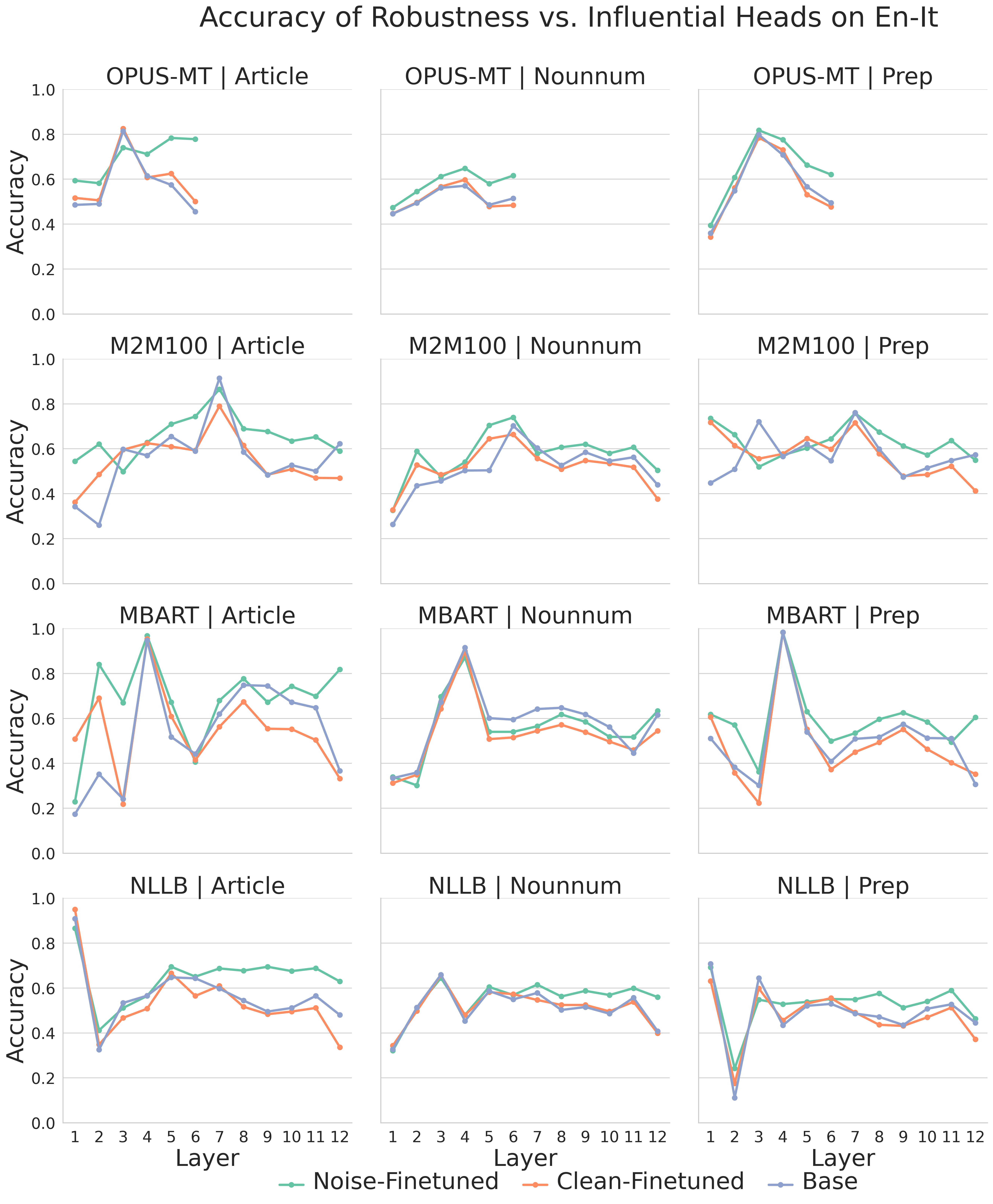}
        \label{fig:influential_robustness_heads_en_it}
    }
    \hfill
    \subfigure[En-Nl]{%
        \includegraphics[width=0.48\textwidth]{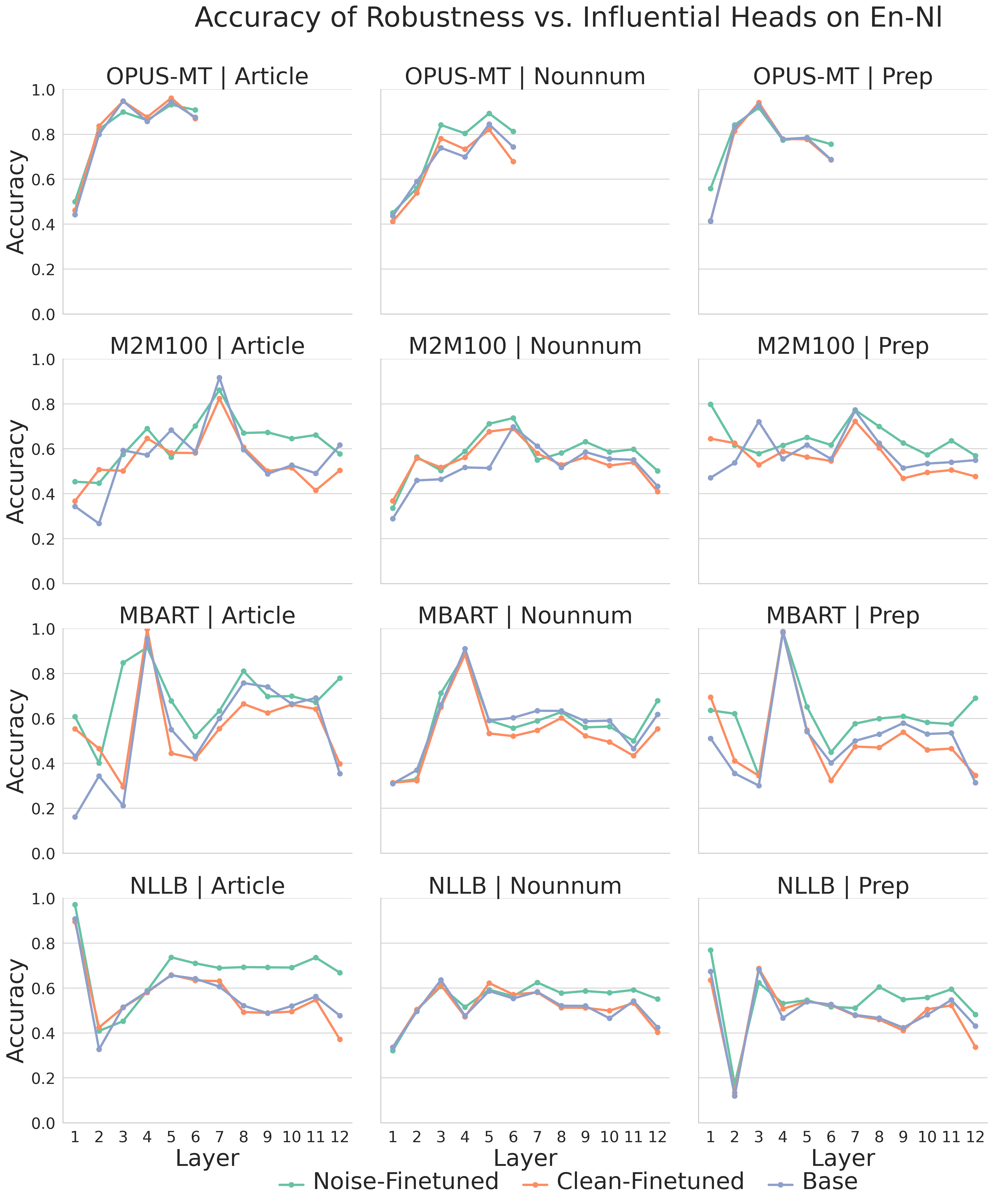}
        \label{fig:influential_robustness_heads_en_nl}
    }
    \caption{Accuracy of Robustness and Influential heads on Fr-Es, En-De, En-It and En-Nl. We find the accuracy is higher for Noise-Finetuned models especially in deep layers.}
    \label{fig:influential_robustness_heads_appendix}
\end{figure*}

Figure \ref{fig:influential_robustness_heads_appendix} shows the accuracy of Robustness and Influential heads on Fr-Es, En-De, En-It and En-Nl.

\subsection{Generalization to Morpheus}
\label{morpheus_results}
To validate the generalization of our analysis, we present the results of using Morpheus. We find similar results in terms of fine-tuning performance, probing, representational distance and similarity between robustness and influential heads (Shown in Table \ref{table:finetuning_results_morpheus} and Figures \ref{fig:morpheus_ged_probing}, \ref{fig:morpheus_cka_distance} and \ref{fig:morpheus_influential_vs_robustness} respectively). We do not present the results of attention to POS tags because the interpretation of results requires granular errors.

\subsubsection{Fine-tuning Results}
We present the results of fine-tuning for robustness to \textit{Morpheus} errors in Table \ref{morpheus_results}.
\subsubsection{GED Probing}
\begin{figure*}
    \centering
    \includegraphics[width=0.95\textwidth,height=0.4\textheight]{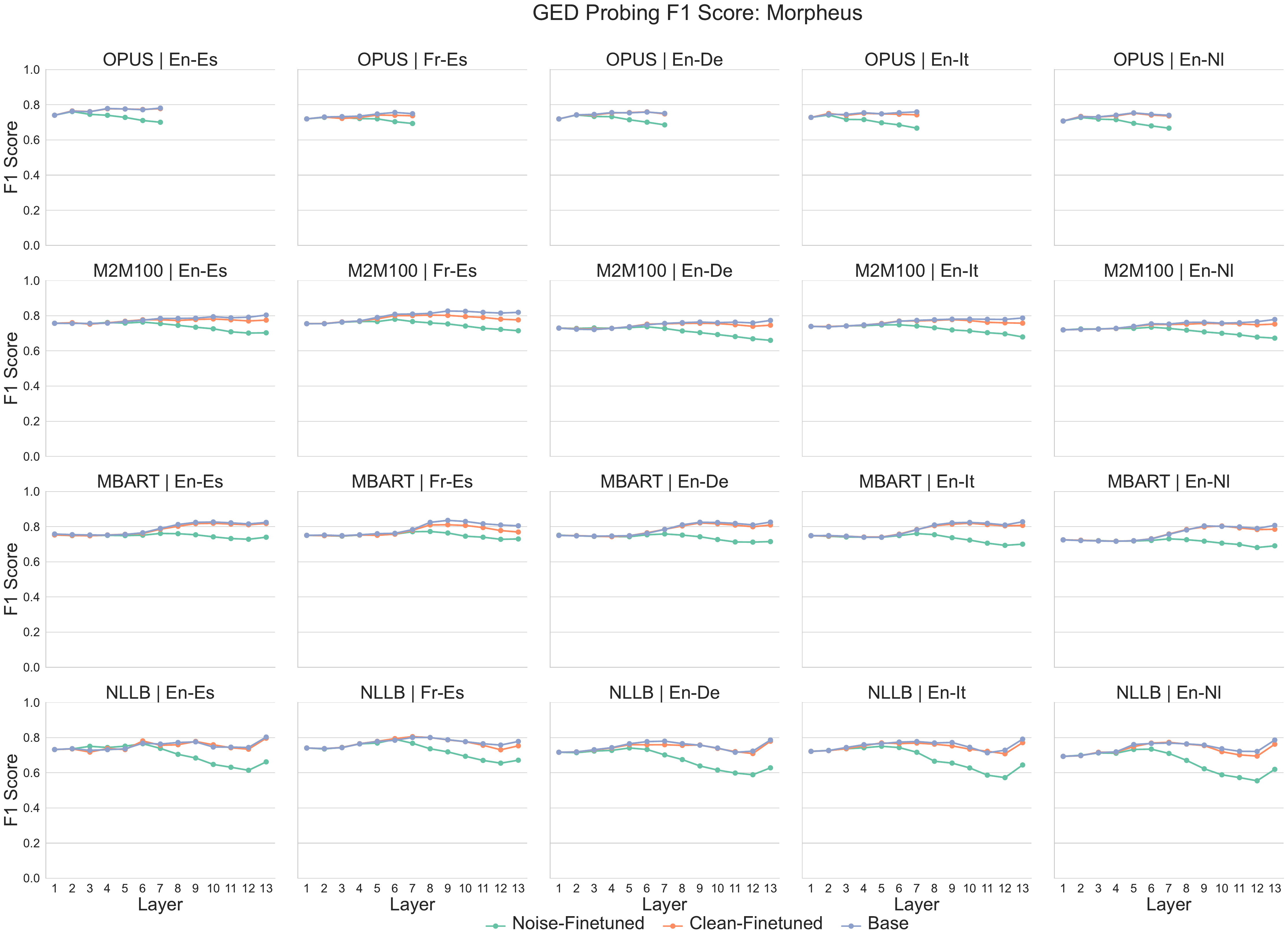}
    \caption{GED probing performance of Noise-Finetuned, Clean-Finetuned and Base models on En-Es, Fr-Es, En-De, En-It and En-Nl on Morpheus errors. GED probing performance of Noise-Finetuned models witnesses a degradation in deeper layers.}
    \label{fig:morpheus_ged_probing}
\end{figure*}

Figure \ref{fig:morpheus_ged_probing} shows the GED probing performance of \textit{Morpheus} errors.
\subsubsection{Representation Distance}
\begin{figure*}
    \centering
    \includegraphics[width=0.95\textwidth,height=0.4\textheight]{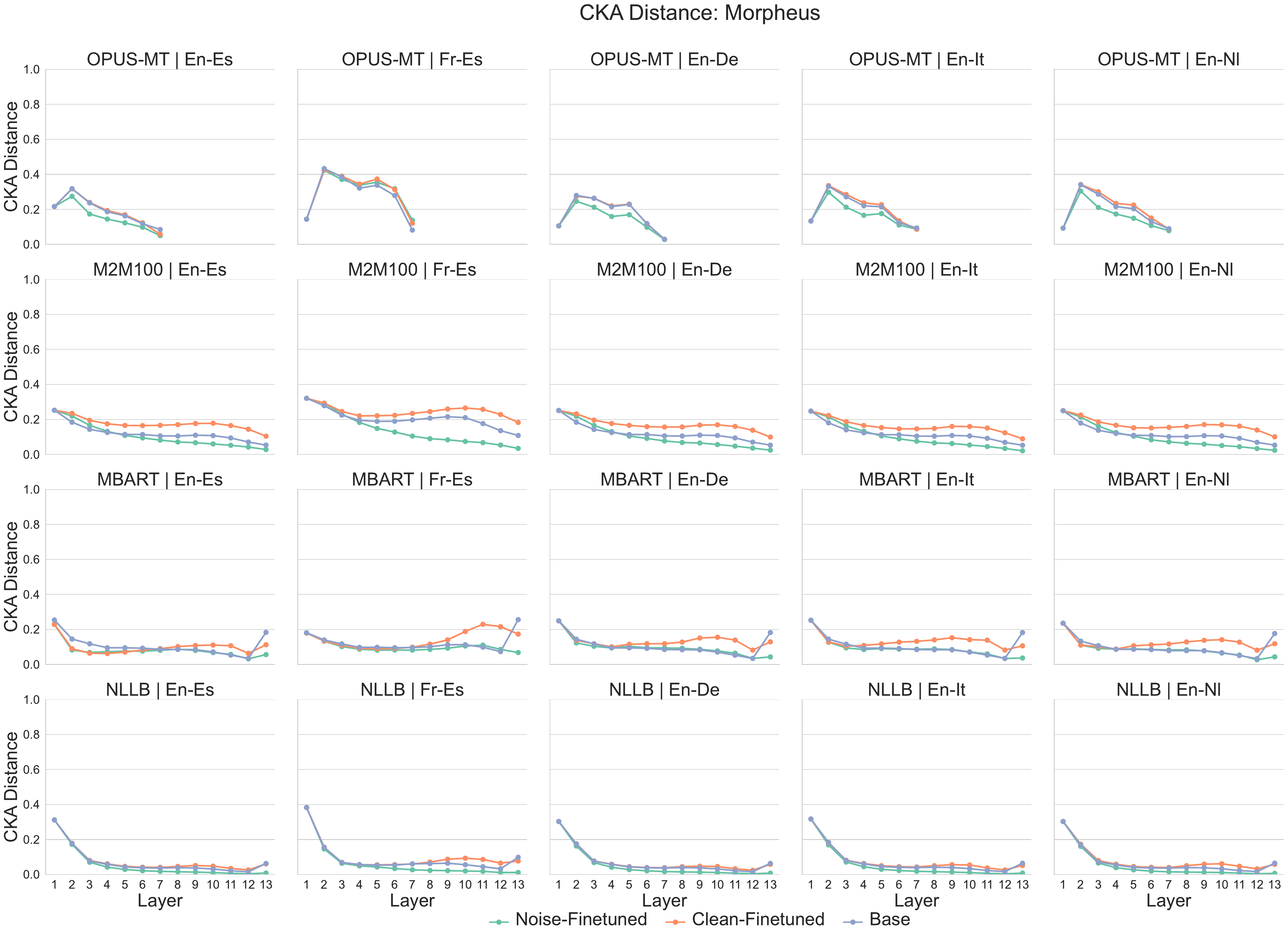}
    \caption{CKA distance of clean and noise word representations across models and errors on En-Es, Fr-Es, En-De, En-It and En-Nl on Morpheus Errors. Noise-Finetuned models drive the representations of the noisy word to be more similar to the clean word.}
    \label{fig:morpheus_cka_distance}
\end{figure*}

Figures \ref{fig:morpheus_cka_distance} shows the CKA distance of clean and noisy word representations of \textit{Morpheus} errors.
\subsubsection{Similarity between Robustness and Influential Heads}
\begin{figure*}
    \centering
    \includegraphics[width=0.95\textwidth,height=0.4\textheight]{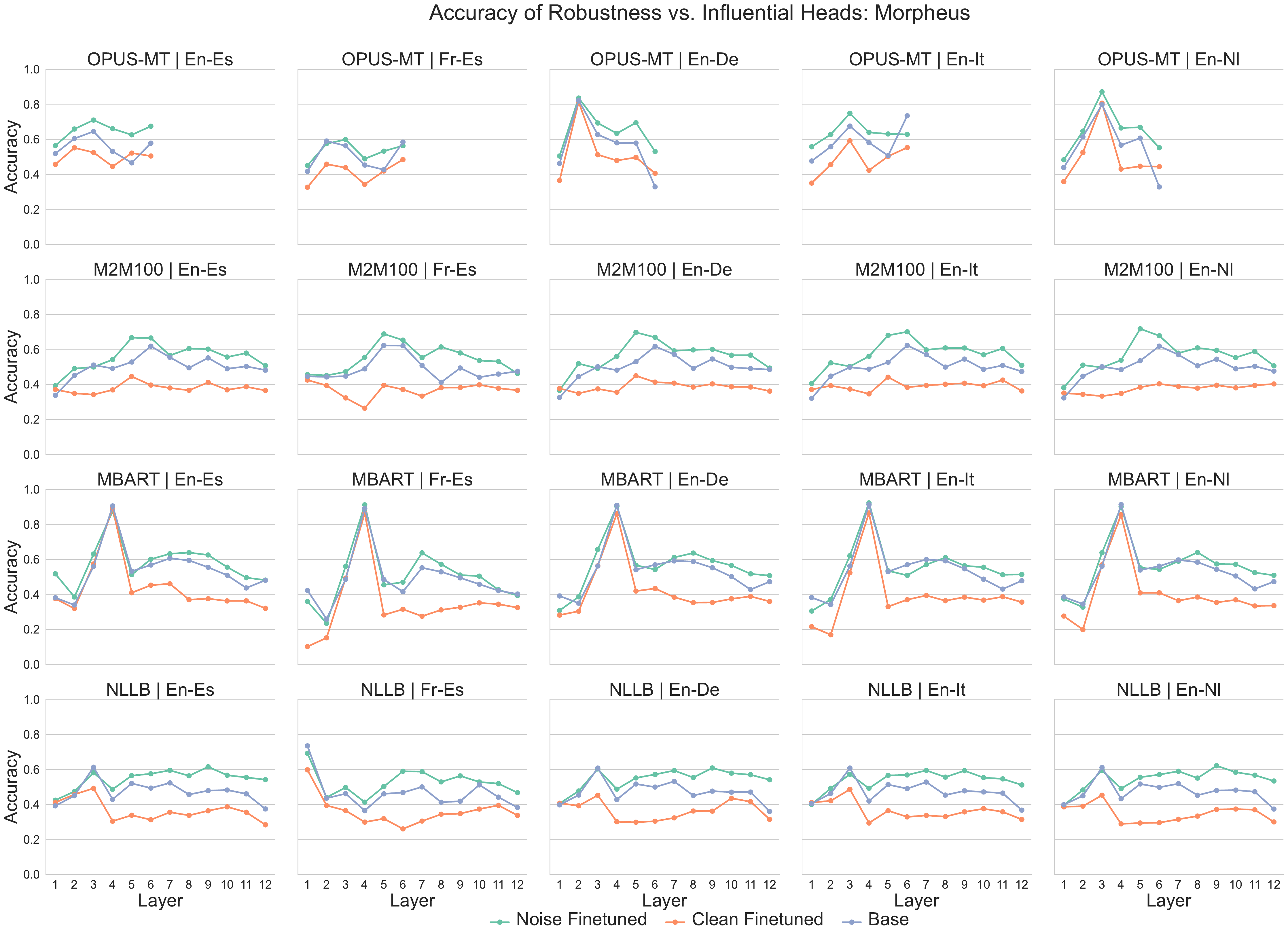}
    \caption{Accuracy of Robustness and Influential heads on En-Es, Fr-Es, En-De, En-It and En-Nl on Morpheus errors. We find the accuracy is higher for Noise-Finetuned models especially in deep layers.}
    \label{fig:morpheus_influential_vs_robustness}
\end{figure*}

Figure \ref{fig:morpheus_influential_vs_robustness} shows the accuracy of Robustness and Influential heads on \textit{Morpheus} errors. 

\clearpage

\begin{table*}[tbh]
\scriptsize
\centering
\begin{tabular}{llrrr}
\midrule
& & \multicolumn{3}{c}{\textbf{Morpheus}} \\
\textbf{Direction} & \textbf{Model} & \textbf{Clean} & \textbf{Noisy} & \textbf{$\Delta$} \\
\midrule
\multirow{13}{*}{En-Es}   & opus-base & 76.03 & 71.78 & 4.25 \\ 
& opus-clean & \textbf{76.2} & 72.28 & 3.92 \\ 
& opus-noisy & 75.85 & \textbf{75.35} & \textbf{0.5} \\ 
\cmidrule{2-5}
& m2m100-base & 75.98 & 71.8 & 4.18 \\ 
& m2m100-clean & \textbf{76.48} & 72.54 & 3.94 \\ 
& m2m100-noisy & 75.79 & \textbf{75.58} & \textbf{0.22} \\ 
\cmidrule{2-5}
& mbart-base & 76.44 & 72.75 & 3.69 \\ 
& mbart-clean & \textbf{78.38} & 74.77 & 3.62 \\ 
& mbart-noisy & 78.27 & \textbf{78.08} & \textbf{0.19} \\ 
\cmidrule{2-5}
& nllb-base & 75.18 & 71.78 & 3.4 \\ 
& nllb-clean & \textbf{75.83} & 72.46 & 3.36 \\ 
& nllb-noisy & 75.64 & \textbf{75.45} & \textbf{0.19} \\ 
\midrule
\multirow{13}{*}{Fr-Es} & opus-base & 71.18 & 67.63 & 3.55 \\ 
& opus-clean & \textbf{73.97} & 70.71 & 3.26 \\ 
& opus-noisy & 73.57 & \textbf{73.05} & \textbf{0.52} \\ 
\cmidrule{2-5}
& m2m100-base & 73.0 & 68.64 & 4.35 \\ 
& m2m100-clean & 73.3 & 70.2 & 3.09 \\ 
& m2m100-noisy & \textbf{73.42} & \textbf{73.23} & \textbf{0.19} \\ 
\cmidrule{2-5}
& mbart-base & 65.36 & 60.31 & 5.06 \\ 
& mbart-clean & 73.07 & 69.69 & 3.39 \\ 
& mbart-noisy & \textbf{74.12} & \textbf{73.62} & \textbf{0.5} \\ 
\cmidrule{2-5}
& nllb-base & 71.34 & 68.53 & 2.81 \\ 
& nllb-clean & 73.1 & 70.68 & 2.42 \\ 
& nllb-noisy & \textbf{73.74} & \textbf{73.43} & \textbf{0.3} \\ 
\midrule
\multirow{13}{*}{En-De} & opus-base & 68.62 & 63.91 & 4.71 \\ 
& opus-clean & \textbf{70.1} & 65.29 & 4.81 \\ 
& opus-noisy & 69.15 & \textbf{68.4} & \textbf{0.75} \\ 
\cmidrule{2-5}
& m2m100-base & 72.92 & 69.02 & 3.9 \\ 
& m2m100-clean & \textbf{74.53} & 71.3 & 3.23 \\ 
& m2m100-noisy & 73.98 & \textbf{73.65} & \textbf{0.33} \\ 
\cmidrule{2-5}
& mbart-base & 74.52 & 69.94 & 4.57 \\ 
& mbart-clean & \textbf{77.38} & 74.1 & 3.28 \\ 
& mbart-noisy & 77.02 & \textbf{76.82} & \textbf{0.2} \\ 
\cmidrule{2-5}
& nllb-base & 65.19 & 62.09 & 3.1 \\ 
& nllb-clean & 66.24 & 63.45 & 2.78 \\ 
& nllb-noisy & \textbf{66.52} & \textbf{66.02} & \textbf{0.5} \\ 
\midrule
\multirow{13}{*}{En-It} & opus-base & 74.82 & 70.14 & 4.69 \\ 
& opus-clean & 76.16 & 72.11 & 4.05 \\ 
& opus-noisy & \textbf{76.22} & \textbf{75.73} & \textbf{0.48} \\ 
\cmidrule{2-5}
& m2m100-base & 74.94 & 70.97 & 3.98 \\ 
& m2m100-clean & \textbf{76.69} & 73.38 & 3.31 \\ 
& m2m100-noisy & 76.59 & \textbf{76.46} & \textbf{0.13} \\ 
\cmidrule{2-5}
& mbart-base & 75.22 & 72.01 & 3.22 \\ 
& mbart-clean & 77.55 & 74.2 & 3.35 \\ 
& mbart-noisy & \textbf{77.58} & \textbf{77.37} & \textbf{0.22} \\ 
\cmidrule{2-5}
& nllb-base & 74.84 & 71.68 & 3.17 \\ 
& nllb-clean & \textbf{75.78} & 72.82 & 2.96 \\ 
& nllb-noisy & 75.59 & \textbf{75.26} & \textbf{0.33} \\ 
\midrule
\multirow{13}{*}{En-Nl} & opus-base & 74.07 & 69.39 & 4.68 \\ 
& opus-clean & 75.32 & 71.5 & 3.82 \\ 
& opus-noisy & \textbf{76.08} & \textbf{75.51} & \textbf{0.58} \\ 
\cmidrule{2-5}
& m2m100-base & 74.84 & 71.45 & 3.4 \\ 
& m2m100-clean & \textbf{75.28} & 71.99 & 3.29 \\ 
& m2m100-noisy & 74.78 & \textbf{74.63} & \textbf{0.15} \\ 
\cmidrule{2-5}
& mbart-base & 74.09 & 71.46 & 2.63 \\ 
& mbart-clean & 77.25 & 74.4 & 2.85 \\ 
& mbart-noisy & \textbf{77.38} & \textbf{77.28} & \textbf{0.1} \\ 
\cmidrule{2-5}
& nllb-base & 72.61 & 69.73 & 2.87 \\ 
& nllb-clean & \textbf{73.64} & 71.13 & 2.51 \\ 
& nllb-noisy & 73.53 & \textbf{73.39} & \textbf{0.15} \\ 

\bottomrule
\end{tabular}
\caption{COMET scores for Morpheus on En-Es, Fr-Es, En-De, En-It and En-NL. The Base model is the original model, Clean is fine-tuned on the clean version of the data, and Noise is fine-tuned on the noisy version (with the same noise as the one they are tested on). We present the performance on the clean and noisy test sets and their difference ($\Delta$).}
\label{table:finetuning_results_morpheus}
\end{table*}

\end{document}